\documentclass[10pt,journal,compsoc]{IEEEtran}

\usepackage{graphicx}
\usepackage{footnote}
\usepackage{diagbox}
\usepackage{multirow} 
\usepackage{threeparttable}
\usepackage{float}
\usepackage{tablefootnote}
\usepackage{hyperref}
\begin{document}

\title{An Overview of Facial Micro-Expression Analysis: Data, Methodology and Challenge}

\author{Hong-Xia~Xie,
        Ling~Lo,
        Hong-Han~Shuai,
        and~Wen-Huang~Cheng,~\IEEEmembership{Senior~Member,~IEEE}
\IEEEcompsocitemizethanks{\IEEEcompsocthanksitem H.-X.~Xie, L.~Lo are with the Institute of 
of Electronics, National Chiao Tung Univresity, Hsinchu, 300 Taiwan.
\\E-mai: \{hongxiaxie.ee08g, lynn97.ee08g\}@nctu.edu.tw. \protect\\ 

\IEEEcompsocthanksitem H.-H.~Shuai is with the Department of Electrical and Computer Engineering, National Chiao Tung Univresity, Hsinchu, 300 Taiwan. \\E-mail: hhshuai@nctu.edu.tw.\protect\\

\IEEEcompsocthanksitem W.-H.~Cheng is with the Institute of Electronics, National Chiao Tung Univresity, Hsinchu, 300 Taiwan, and the Artificial Intelligence and Data Science Program, National Chung Hsing University, Taichung, 400 Taiwan. \\E-mail: whcheng@nctu.edu.tw.
}
}

\IEEEtitleabstractindextext{%
\begin{abstract}
Facial micro-expressions indicate brief and subtle facial movements that appear during emotional communication. In comparison to macro-expressions, micro-expressions are more challenging to be analyzed due to the short span of time and the fine-grained changes. In recent years, micro-expression recognition (MER) has drawn much attention because it can benefit a wide range of applications, e.g. police interrogation, clinical diagnosis, depression analysis, and business negotiation. In this survey, we offer a fresh overview to discuss new research directions and challenges these days for MER tasks. For example, we review MER approaches from three novel aspects: macro-to-micro adaptation, recognition based on key apex frames, and recognition based on facial action units. Moreover, to mitigate the problem of limited and biased ME data, synthetic data generation is surveyed for the diversity enrichment of micro-expression data. Since micro-expression spotting can boost micro-expression analysis, the \textit{state-of-the-art} spotting works are also introduced in this paper. At last, we discuss the challenges in MER research and provide potential solutions as well as possible directions for further investigation.

\end{abstract}

\begin{IEEEkeywords}
Facial Micro-expression, Recognition, Spotting, Action Units, Deep Learning, Survey.
\end{IEEEkeywords}}

\maketitle

\IEEEdisplaynontitleabstractindextext

%
\IEEEpeerreviewmaketitle

\IEEEraisesectionheading{\section{Introduction}\label{sec:introduction}}
\IEEEPARstart{F}{acial} micro-expression (ME) is a result of \textit{conscious suppression} (intentional) or \textit{unconscious repression} (unintentional), which can be viewed as a "leakage" of people's true feelings~\cite{ekman2009telling}. MEs are brief involuntary facial expressions that usually appear when people are trying to conceal their true feelings, especially in high-stake situations. Hence, micro-expression recognition (MER) research enables greater awareness and sensitivity to subtle facial behaviors, and is an important subject for human emotion and affective phenomena understanding, which has been explored by various disciplines such as psychology, sociology, neuroscience, computer vision, etc. Such skills are useful for psychotherapists, interviewers, and anyone working in communications. 
 
 
\begin{figure}[!t]
\centering
\includegraphics[scale = 0.61]{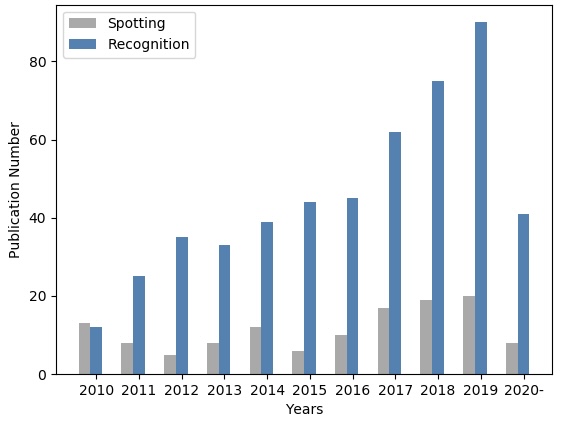}
\caption{The publication number of Micro-expression Recognition and Spotting from 2010 to November 2020 (Source: Web of Science).}
\label{fig:publications}
\end{figure} 

\begin{figure*}[!t]
\centering
\includegraphics[scale = 0.3]{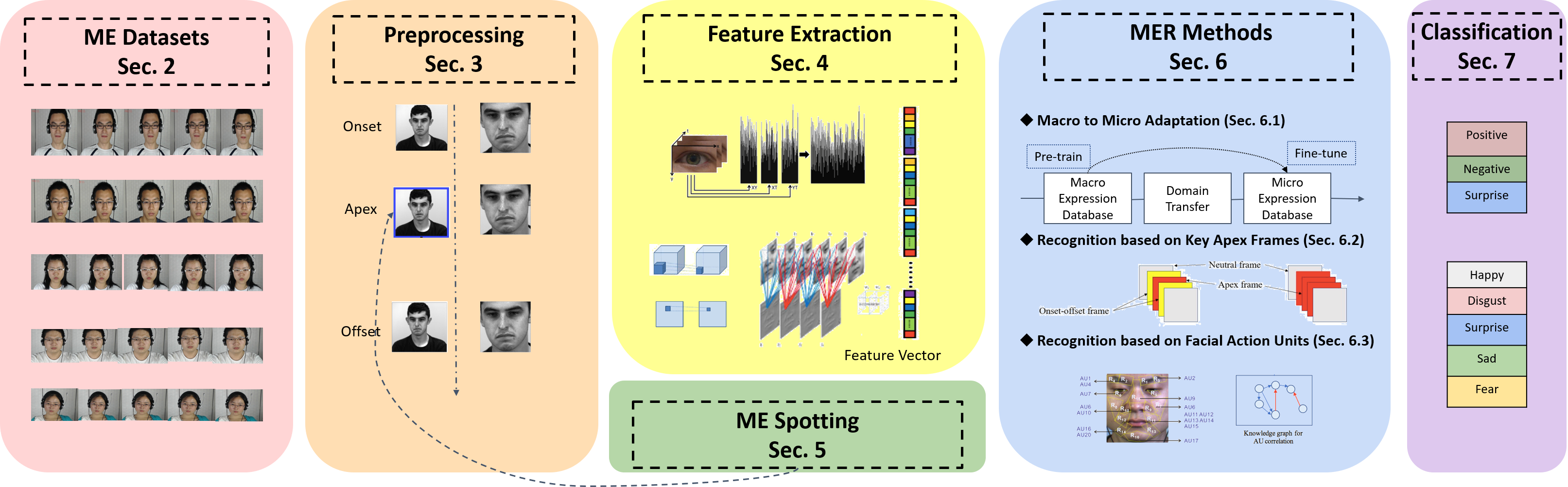}
\caption{The organization of this survey is structured according to the general MER pipeline.}
\label{fig:overall}
\end{figure*} 
 
\begin{table*}[!t]
\renewcommand{\arraystretch}{1.3}
\caption{Academic challenges for micro-expression recognition and spotting.}
\label{challenges}
\centering
\begin{tabular}{|c||c||c||c||c||c| }
\hline
Challenge &Year&Dataset&Task&Evaluation Metric&Event \\
\hline\hline
MEGC2018~\cite{MEGC2018}&2018&CASME II, SAMM&Cross-database (HDE, CDE)&UAR, WAR, F1-score& FG$^1$ \\
\hline 
MEGC2019~\cite{MEGC2019}&2019&CAS(ME)$^2$, SAMM&Recognition, Spotting (CDE)&UAR, WAR, F1-score& FG$^1$ \\
\hline 
MEGC2020~\cite{li2020megc2020}&2020&CAS(ME)$^2$, SAMM Long Video&Spotting&F1-score& FG$^1$ \\
\hline
MER2020*&2020&Synthetic Data&Recognition&F1-score& ICIP$^2$ \\
\hline
\hline
\end{tabular}
\begin{tablenotes} 
\item $^{*}$ MER2020: \url{http://mer2020.tech/}
\item $^{1}$ FG: IEEE International Conference on
Automatic Face and Gesture Recognition
\item $^{2}$ ICIP: IEEE International Conference on Image Processing
\end{tablenotes} 
\end{table*}

\subsection{The Difference between Macro-expression and Micro-expression}
Micro-expressions occur when people are trying to conceal or repress their true feelings. On the contrary, macro-expressions are easy to be perceived in daily interactions. The major difference between macro-expressions and micro-expressions lies in their duration. Although there is no strict rule of the threshold to distinguish one from the other, most agreed that macro-expression usually last from 0.5 to 4 seconds while micro-expressions should be no longer than 0.5 seconds~\cite{shen2012duration}. Most researches \cite{yap2019samm, yan2013casme, yan2014CASMEII} set 0.5 seconds as the threshold, while 0.2 seconds duration is also regarded as a boundary for differentiating micro- and macro-expressions as validated in~\cite{shen2016electrophysiological}. Besides, the neural mechanisms underlying the recognition of micro-expression and macro-expression are different, showing different electroencephalogram (EEG) and event-related potentials (ERPs) characteristics. The brain regions responsible for their differences might be the inferior temporal gyrus and the frontal lobe~\cite{shen2016electrophysiological}. General speaking, the macro-expression shows higher intensity and visibility when compared to the micro-expression.

\subsection{Facial Action Coding System}
Facial Action Coding System (FACS) was first proposed by Ekman and Friesen and updated in 2002~\cite{ekman1997face,rosenberg2020face} to factorize the composition of micro-expressions. It is the most widely used coding scheme for decomposing facial expressions into individual muscle movements, called Action Units (AUs). With FACS, every possible facial expression can be described as a combination of AUs. There are 32 facial muscle-related actions, and 6 extra unspecific miscellaneous Action Descriptors (ADs)~\cite{wang2015micro}. 

Yet, the professional training of FACS encoding experts is time-consuming. Professional encoders usually need to receive 100 hours of training, and in practice the encoding process takes 2 hours to encode a 1-minute video on average~\cite{pantic2009machine}. Thus, an automatic MER system with high accuracy can be very helpful and valuable.

Since AUs are descriptive for certain facial configurations, specific systems are proposed to explore the relationship between facial muscle movements (AUs) and human emotions, e.g., Emotional Facial
Action Coding System (EMFACS-7)~\cite{EMFACS7}, Facial Action Coding System Affect Interpretation Dictionary (FACSAID)~\cite{ekman2002facial} and the System for Identifying Affect Expressions by Holistic Judgments (Affex)~\cite{FACSAID}. It can be found that various mapping strategies for AUs and emotions are adopted by existing facial expression datasets due to the lack of standard guidelines~\cite{duran2017coherence}. 

\subsection{Academic Challenges for MER}
Academic challenge is a competition created by academic experts for leveraging the power of open innovation to advance the \textit{state-of-the-art} in a particular field. Within the computer vision domain, several academic challenges related to micro-expression recognition and spotting have been proposed. Some known events in recent years are summarized in Table~\ref{challenges}.


\makesavenoteenv{tabular}
\begin{table*}[!t]
\begin{threeparttable} 
\renewcommand{\arraystretch}{1.3}
\caption{A list of public datasets on spontaneous micro-expression.}
\label{Micro-Expression Dataset}
\begin{center}
\item
\begin{tabular}{|c||c||c||c||c||c||c||c||c||c| }
\hline
 Dataset& Macro/Micro& Videos& FPS & Resolution& FACS & Emotion& Subjects&AU&Index* \\
\hline\hline
CASME & micro& 195 & 60& 640$\times$480,1280$\times$720& Yes& 8&19&No&On,Apex,Off\\
\hline
\qquad\quad HS & & 164 & 100&  & & &16&&On,Off\\
SMIC VIS & micro& 71 & 25& 640$\times$480 & No&3&8&No&N/A\\
\qquad\quad NIR & & 71 & 25&  & & &8&&N/A\\
\hline
CASME II & micro& 247 & 200& 640$\times$480 & Yes& 5&26&Yes&On,Apex,Off\\
\hline
SAMM & micro& 159 & 200& 2040$\times$1088 & Yes& 7&32&Yes&On,Apex,Off\\
\hline
CAS(ME)$^2$ & macro \& micro& 300 \& 57 & 30& 640$\times$480  & No& 4&22&N/A&On,Apex,Off\\
\hline
SAMM Long Videos & macro \& micro& 343 \& 159 & 200& 2048$\times$1088 & Yes& N/A&30&Yes&On,Apex,Off\\
\hline
\end{tabular}
\end{center}
\begin{tablenotes}
\small
\item $^{*}$ On, Apex, Off: Onset frame, Apex frame, Offset frame, respectively.
\end{tablenotes} 
\end{threeparttable} 
\end{table*}

\subsection{Outline of this Paper}

Numerous works have been accomplished to mitigate the challenging MER task. The publication counting from 2010 to November 2020 is shown in Fig.~\ref{fig:publications}. Several surveys on MER have been published in recent years~\cite{yan2014micro,li2017towards,goh2018micro,oh2018survey,merghani2018review,takalkar2018survey,zhou2020survey}. However, most of them have been focused on traditional image-processing methods and a fresh overview to discuss new directions and challenges the MER faces today is necessary.
In this survey paper, therefore, we provide a more comprehensive and in-depth study of existing approaches, e.g. including over 100 papers for MER until November 2020 which have not been reviewed in the previous survey papers. Based on the nature of MEs, e.g., the short span of time and the fine-grained changes, we introduce MER approaches from three novel aspects: macro-to-micro adaptation, recognition based on key apex frames, and recognition based on facial action units. Moreover, to mitigate the problem of limited and biased ME data, synthetic data generation algorithms are surveyed for the diversity enrichment of ME data. To the best of our knowledge, this is the first review to summarize the synthetic data generation solutions based on spontaneous ME datasets.
Since micro-expression spotting can boost micro-expression analysis, the \textit{state-of-the-art} spotting works are also introduced. At last, we discuss open challenges and provide potential solutions as well as future research directions.

The whole systematic framework of MER is present in Fig.~\ref{fig:overall}. The outline of this paper adheres as follows: Section~\ref{sec: Datasets} introduces ME data collection, current public ME datasets, and popular data synthesis methods. Section~\ref{sec:Preprocessing} describes pre-processing techniques commonly used for ME data. Section~\ref{sec: Feature Representation} provides a detailed review of feature extraction methods based on handcrafted features and deep learning-based methods. Discussions of ME spotting are present in Section~\ref{sec:ME spotting}. Section~\ref{sec:ME approches} summarizes the commonly-used techniques on MER recently. The loss function widely used for MER is introduced in Section~\ref{sec: Loss Function Design}. Section \ref{sec: Overall Comparison/discussion} presents the overall performance comparison. Open MER challenges and the possible solutions are arranged in Section~\ref{sec: Challenges} and some potential future directions for research are suggested in Section~\ref{sec: Future Recommendation}.

\section{Datasets}\label{sec: Datasets}
\subsection{ME Data Collection}
\textbf{Emotion Classes.} In discrete emotion theory, there are many different basic emotion definition systems, of which the most popular one adopted in the computer vision community is conducted by Ekman and Friese~\cite{ekman1992_emotionclass}, where the basic emotions are divided into six categories of anger, disgust, fear, happiness, sadness, and surprise. An extra emotion of contempt is added in later researches.

\begin{figure}[!t]
\centering
\includegraphics[scale = 0.33]{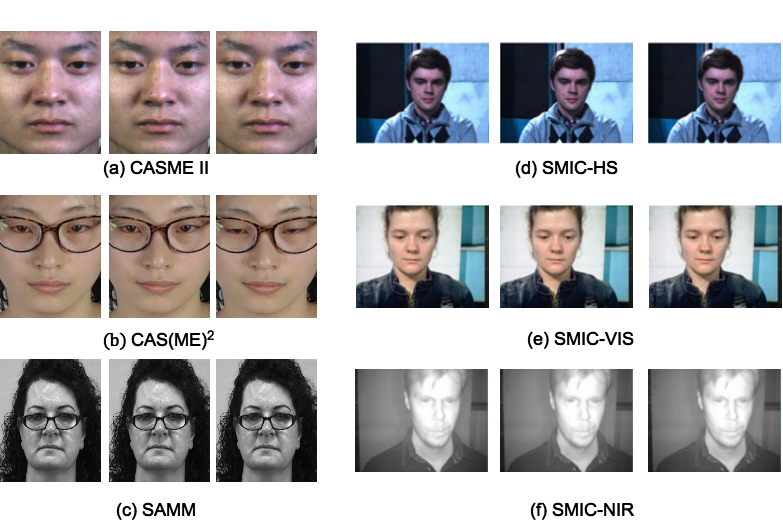}
\caption{Sample frames from different ME datasets.}
\label{fig:ME_datasets}
\end{figure}

Current public ME datasets divide micro-expression emotions into different classes according to their collection strategy. The emotion labels should take into account AUs, participants' self-report as well as the stimuli video content, etc. To reduce unpredictability and bias of emotion classes, 
Spontaneous Micro-expression Corpus (SMIC)~\cite{SMIC} group emotion categories from CASME II and SAMM~\cite{yap2019samm} into three main classes (positive, negative, surprise)~\cite{davison2018objective} based on the AUs with FACS coding. Fig.~\ref{fig:ME_datasets} shows sample frames from current ME datasets.


\subsubsection{Apparatus Setup}
Since the frame rate and resolution could affect the performance of MER~\cite{merghani2019implication}, existing ME datasets are usually collected in a strictly lab-controlled environment. The observers are kept out of sight to maximise the chances of natural suppression by making participants as comfortable as possible. 
The participants are asked to watch video clips in front of a screen and avoid any body movement, which can help to exclude habitual movements.

\subsubsection{Emotional stimuli} 
Spotaneous ME datasets were usually elicited by emotional stimuli, e.g., images, movies, music~\cite{coan2007-EmotionElicitation}. MEs are more likely to occur under high-arousal stimuli, so video clips with high emotional valence are proved to be effective materials for eliciting MEs~\cite{SMIC}.




\textbf{Emotion Label Rating Quantification.} To explicitly quantify the personal emotional response to the emotional stimulus, a predefined scale that a subject can assign to the perceived emotion response is usually needed. SMIC~\cite{SMIC}, CASME~\cite{yan2013casme}, and CASME II~\cite{yan2014CASMEII} databases assigned an emotion label to stimuli videos based on self-reports completed by participants. Relatively, SAMM adopted Self-Assessment Manikins (SAM)~\cite{lang1980behavioral_SAM}, which contains the valence, dominance, and arousal judgments in rating emotional response. 

\subsubsection{Micro-expression Inducement}
\textbf{Surppressing}. To evoke MEs, there must be enough pressure to motivate participants to conceal their true feelings. SMIC, CASME, and SAMM asked participants to fully suppress facial movements during the whole experiment so that MEs may occur. For CASME II, half participants were asked to keep neutral faces when watching video clips while other participants were enforced to suppress the facial movements when they realize there is a facial expression.


\textbf{Not All Subjects Show Micro-expressions.} According to Ekman’s research, when people are telling lies, about half of them might show MEs, while the other half do not~\cite{ekman2009lie}. The reason why only some people show MEs is still unclear. In SMIC, four participants did not show any ME at all throughout the course of 35-minutes video watching.


\subsection{Spontaneous Micro-Expression Dataset}
Since posed ME datasets are collected by intentionally controlled, it contradicts the natural occurrence of MEs. In this paper, we only focus on \textbf{spontaneous ME datasets}. Table~\ref{Micro-Expression Dataset} provides an overview of spontaneous ME datasets.

\textbf{Chinese Academy of Sciences Micro-Expression (CASME)}~\cite{yan2013casme} dataset consists of 195 video samples from 19 valid subjects with a frame rate of 60 fps. The samples for 8 emotions are highly imbalanced (5 happiness, 6 sadness, 88 disgust, 20 surprise, 3 contempt, 2 fear, 40 repression and 28 tense). Notably, in CASME, each sample is recorded with two different cameras and environmental settings, namely Class A and Class B. Class A was recorded by BenQ M31 camera with natural light, while class B was by GRAS-03K2C camera and two LED lights.

\textbf{Chinese Academy of Sciences Micro-Expression II (CASME II)}~\cite{yan2014CASMEII} is an improved version of CASME collected in a well-controlled lab environment. It contains 247 ME sequences from 26 subjects with 7 categories, including happiness, disgust, surprise, fear, sadness, repression and others, which were labeled based on AUs, participants’ self-report and the content of stimuli videos. All subjects are Chinese.

\begin{figure}[!t]
\centering
\includegraphics[scale = 0.33]{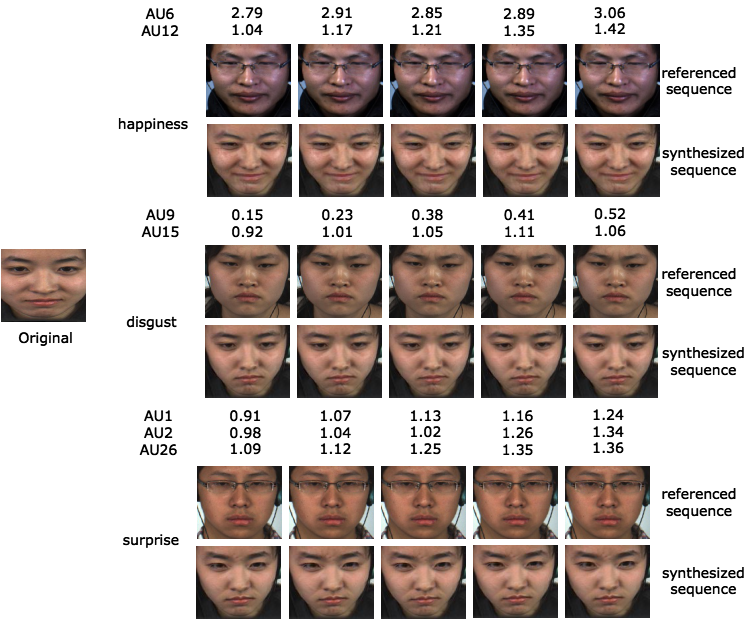}
\caption{Synthetic faces generated by AU-ICGAN from CASME II dataset. The number above each image indicates the target AU intensities of chosen AUs~\cite{xie2020assisted}.}
\label{fig:synthetic}
\end{figure}

\textbf{Spontaneous Micro-expression Corpus (SMIC)}~\cite{SMIC} is composed of 164 ME sequences from 16 subjects filmed at 100 fps. It is one of the first to include spontaneous MEs through emotional inducement experiments. Most subjects are Asians, and some are from other ethnicity. Ethnicity is more diverse than previous datasets with ten Asians, nine Caucasians and one African participant.
For data reliability, ME clips in SMIC are classified into three classes: positive, negative and surprise. While positive (happy) and surprise only include one target emotion each, negative includes four emotions (sad, anger, fear and disgust).

\textbf{Spontaneous Actions and Micro-Movements (SAMM)}~\cite{yap2019samm} has 159 ME sequences from 29 subjects. The seven classes include contempt, disgust, fear, anger, sadness, happiness and surprise.
The ethnicities of participants are diverse and the gender split is even.

\subsection{Macro- and Micro-Expression Datasets}

Considering real-world scenarios, macro- and micro-expressions could co-occur, there are two public datasets combining macro and micro facial expressions, i.e., SAMM Long Videos~\cite{yap2019samm} and CAS(ME)$^2$~\cite{qu2017cas}. Both of them can be empolyed for macro-expression and micro-expression spotting from long videos. 

\textbf{The Chinese Academy of Science Macro- and Micro-expression (CAS(ME)$^2$)} dataset was established by the Chinese Academy of Science. The dataset contains 300 macro-expressions and 57 micro-expressions, with four different emotional labels: positive, negative, surprise and others from 22 participants (13 females and 9 males). The expression samples are coded with the onset, apex, and offset frames, with AUs marked and emotions labeled. 

\textbf{SAMM Long Videos} consists of 147 long videos with 343 macro-expressions and 159 micro-expressions. The frame rate is 200 fps, and 0.5 seconds is set as the threshold for classifying macro- ($\geq$ 0.5 seconds) and micro-expressions (\textless 0.5 seconds). Emotion labels are not provided.

\subsection{Synthetic Data Generation}
While building ME databases, it is not only challenging to trigger an ME but also very difficult to label one. Labeling ME data requires human labor as well as professional domain knowledge. Even with professional training, it is reported that up to only 47\% labeling accuracy can be achieved for a human expert~\cite{frank2009see}. Thus the size of existing databases is usually very limited. Synthetic datasets are introduced in this situation when annotating the ground-truth is a time-consuming work. In general, synthetic data have been successfully used in many areas in computer vision from learning low-level visual features~\cite{freeman2000learning, butler2012naturalistic,dosovitskiy2015flownet} to high-level tasks~\cite{wang2019learning, richter2016playing,papon2015semantic}. Especially, synthesizing facial data is more challenging than other basic objects as the fidelity of human faces is hard to preserve~\cite{9229137}.  Queiroz \textit{et al.}~\cite{queiroz2010generating} presented a methodology for generation of facial ground truth with synthetic faces. The 3D face model database can control face actions as well as illumination conditions, allowing to generate animation with different facial expressions and eye motions with the ground truth of the facial landmark points provided at each frame. The resulting
Virtual Human Faces Database (VHuF) dataset can simulate realistic  skin textures extracted from real photos. Abbasnejad \textit{et al.}~\cite{abbasnejad2017using} established a large-scale dataset of facial expression using a 3D face model. It consists of shape and texture models to create different subjects with different expressions. Their experimental results showed that the synthesized dataset enables efficiently deep network training for expression analysis. 

Apart from synthetic databases generated by 3D morphable models, an alternative way to produce synthetic data is using Generative Adversarial Network (GAN). Zhang \textit{et al.}~\cite{zhang2018joint} took advantage of GAN and proposed a network exploiting different poses and expressions jointly for facial image synthesis and pose-invariant facial expression recognition. The generated face images with different expressions under arbitrary poses can enlarge and enrich expression dataset and benefit the recognition accuracy. Cai \textit{et al.}~\cite{cai2019identity} synthesized images with a same face in different expressions using a conditional generative model. The resulting dataset consists of sets of images and each image set contains a same identity in different synthetic expressions, which benefits the identity-free expression recognition. Different from large-scale facial expression, Xie \textit{et al.}~\cite{xie2020assisted} proposed AU Intensity Controllable Generative Adversarial Networks (AU-ICGAN) for micro-scale expression data synthesis. To enrich the limited data samples, their AU-ICGAN aims to generate face images with specific intensities of action unit to simulate real-world ME data. In addition, the image structure similarity along with image sequence authenticity are taken into consideration so that the generated ME image sequences can be more realistic. Their experimental results show the merits of using an auxiliary synthetic dataset for training deep recognition networks. Fig.~\ref{fig:synthetic} shows the synthetic results for enriching ME datasets.

\begin{figure}[!t]
\centering
\includegraphics[scale = 0.32]{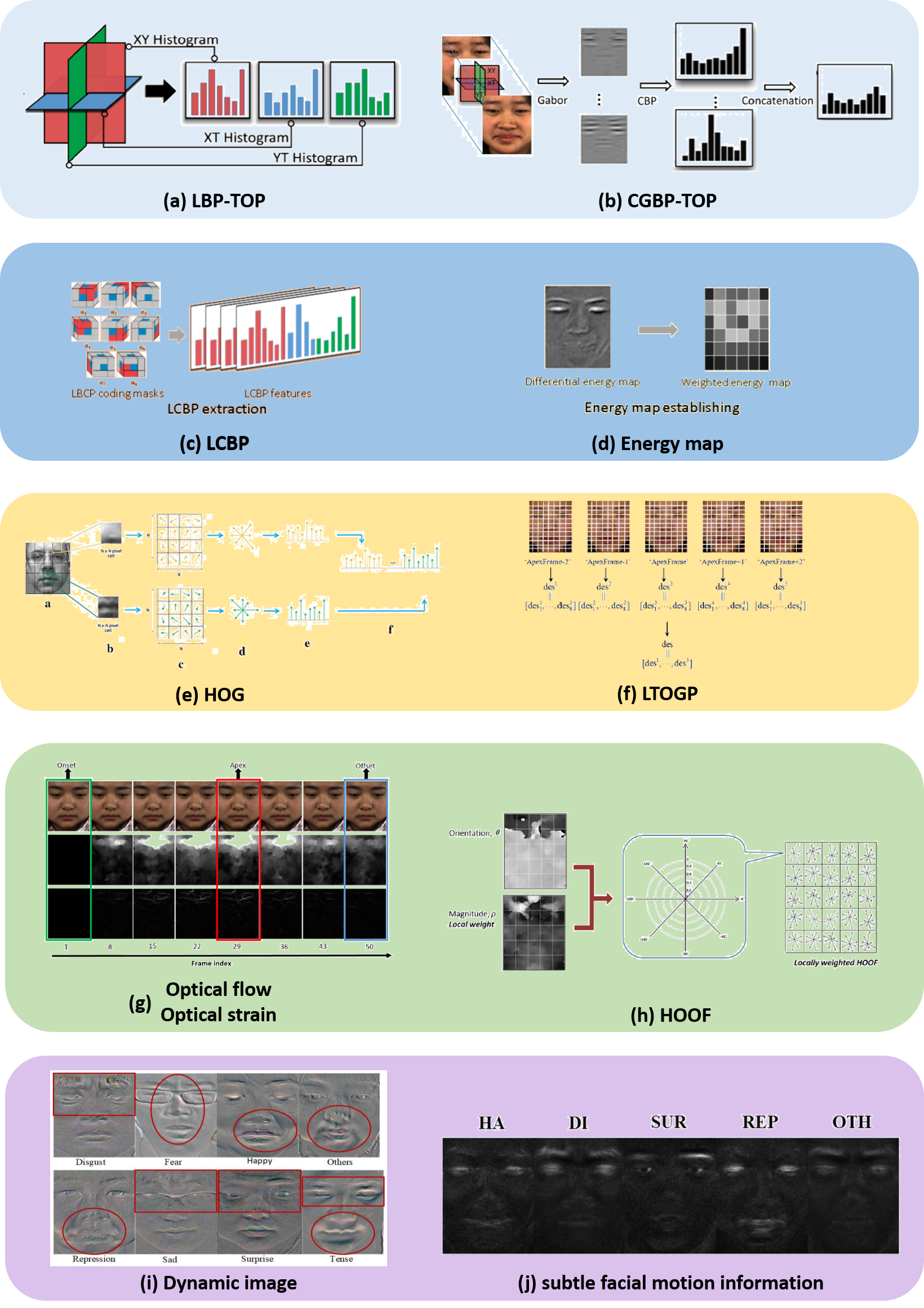}
\caption{An illustration of handcrafted features: (a) LBP-TOP~\cite{yu2019lcbp}, (b) CGBP-TOP~\cite{hu2019gender}, (c) LCBP ~\cite{yu2019lcbp}, (d) energy map~\cite{yu2019lcbp}, (e) HOG~\cite{carcagni2015facial}, (f) LTOGP~\cite{niu2018LTOGP},  (g) optical flow (optical strain)~\cite{liong2018biwoof}, (h) HOOF~\cite{liong2018biwoof}, (i) dynamic image~\cite{sun2020dynamic} and (j) subtle facial motion information~\cite{huang2017discriminative}.}
\label{fig:handcrafted}
\end{figure}

\section{Pre-processing}\label{sec:Preprocessing}
The pre-processing stage in MER consists of all steps required before the extraction of meaningful features can commence. One of the important aim of the pre-processing stage is to detect and align faces into a common reference, so that the features extracted from each face correspond to the same semantic locations. It removes rigid head motion and, to some extent, the anthropomorphic variations among people. After alignment, the subtle micro muscle movements can be further magnified in order to enhance the discriminative characteristic. 

\subsection{Face Detection}
The first step of any face analysis method is to detect the face region. The Viola\&Jones (V\&J) face detector~\cite{viola2001rapid} can give real-time and robust near-frontal face in an image and thus is one of the most popular face detector. Their cascade classifier is based on Haar feature, and its reliability and computational simplicity make it a widely employed approach. Another widely-used face detector is based on Histogram of Gradient (HOG)~\cite{dalal2005histograms,sanchez2017robust}. 
HOG-based face detectors share the merit of computational efficiency with V\&J face detectors, but they also have same short of being unable to deal with non-frontal face images.
With the rapid development of Convolution Neural Network (CNN), more CNN-based face detectors are adopted in popular open source libraries. For example, Max-Margin Object Detection (MMOD)~\cite{king2015max} is used in dlib\footnote{http://dlib.net/}, and a well-trained single-Shot-Multibox detector using ResNet-10 as the backbone is provided in OpenCV\footnote{https://opencv.org/}. Also, the CNN-based face detectors are reported more accurate and robust under occlusions~\cite{masi2018deep}.

\begin{table*}[!t]
\renewcommand{\arraystretch}{1.3}
\caption{Performance comparison among micro-expression spotting methods}
\label{tab: spotting}
\centering
\begin{threeparttable}
\begin{tabular}{|c||c||c||c||c||c||c||c|c|c| }
\hline
\multirow{2}{*}{Methods} &\multirow{2}{*}{Year} &\multirow{2}{*}{Features}&\multirow{2}{*}{Protocol}&\multirow{2}{*}{Datasets} & \multicolumn{3}{c|}{Experimental Results}\\
\cline{6-8}
& & & &  & ACC & F1-score & recall \\
\hline\hline
\multirow{2}{*}{\cite{zhang2020spatio}}&
\multirow{2}{*}{2020}&\multirow{2}{*}{Specific Pattern}&\multirow{2}{*}{-}& SAMM & 0.090 & 0.133 & 0.258  \\
&&&& CAS(ME)$^2$	& 0.029	&0.055	&0.456\\
\hline
\cite{verburg2019micro} &2019&HOOF$\&$RNN &LOSO& SAMM &0.045&0.082	&0.465	\\ 
\hline
\multirow{2}{*}{\cite{li2019spotting}}&
\multirow{2}{*}{2019}&\multirow{2}{*}{LTP-ML}&\multirow{2}{*}{LOSO}& SAMM & 0.017 & 0.032 & 0.296  \\
&&&& CAS(ME)$^2$	& 0.009	&0.018	&0.281\\
\hline

\multirow{2}{*}{\cite{tran2019dense}}&
\multirow{2}{*}{2019}&\multirow{2}{*}{HIGO-TOP $\&$ HOG-TOP}&\multirow{2}{*}{LOSO}& SMIC-VIS-E & - & 0.620 & -  \\
&&&& CASME II	& -	&0.860	&-\\
\hline
\cite{beh2019micro} &2019&HOG&LOSO& CASME II &0.823&-&-	\\ 
\hline
\cite{nag2019facial} &2019&time-constrasted feature&-& CASME II &0.815&-&-	\\ 
\hline
\multirow{2}{*}{\cite{han2018cfd}}&
\multirow{2}{*}{2018}&\multirow{2}{*}{collaborative feature difference (CFD)}&\multirow{2}{*}{-}& SMIC-E & - & - & 0.942(AUC)  \\
&&&& CASME II	& -	&-	&0.971(AUC)\\
\hline
\multirow{2}{*}{\cite{duque2018micro}}&
\multirow{2}{*}{2018}&\multirow{2}{*}{reiesz pyramid}&\multirow{2}{*}{LOSO}& SMIC-HS & - & - & 0.898(AUC)  \\
&&&& CASME II	& -	&-	&0.951(AUC)\\
\hline

\end{tabular}

\end{threeparttable}

\end{table*}

\subsection{Facial Landmark Detection}
While recognizing face regions in images is the vital beginning, aligning faces in a video is even more crucial to be used in most ME recognition frameworks. To align faces in different frames, the coordinates of localized facial landmarks are first detected. Then, the faces in different frames are aligned into a referenced face according to the location of the landmark keypoints. Proper face alignment can improve the recognition accuracy significantly because the subtle motion of faces over time can be captured. 
Active Shape Model (ASM)~\cite{milborrow2008locating} is one of the face models that is frequently used. 
Discriminative response map fitting (DRMF)~\cite{asthana2013robust} is a regression approach with strong ability in general face fitting.
With lower computational consumption and real-time capabilities, the DRMF model can handle occlusions, dynamic backgrounds and various illumination conditions.

In recent years, deep-learning methods have been widely used for facial landmarks localization.
Cascaded-CNN, which predicts the facial landmarks in a cascaded fashion, has become a \textit{state-of-the-art }method for its high accuracy and speed. Tweaked Convolutional Neural Networks (TCNN)~\cite{ wu2017facial} can harness the robustness of CNNs for landmark detection in an appearance-sensitive manner. 

Besides, for facial landmark alignment or face registration, generic method like Kanade-Lucas-Tomasi (KLT)~\cite{birchfield1997derivation} is also proposed to track point features in different frames among a whole video. 

\subsection{Motion Magnification}
Through motion magnification, subtle facial expressions become more recognizable. Local Amplitude-based Eulerian Motion Magnification (EMM)~\cite{le2019seeing} is commonly used to magnify the micro-level movements by exaggerating the differences in the brightness and color of pixels at the same spatial location across two consecutive frames. Compared to local magnification, global-scale Lagrangian Motion Magnification (LMM)~\cite{le2018micro} was proposed for explicitly tracking all the displacements within a video in both the spatial and temporal domains. In LMM, each pixel can be traced back and forth through the timeline. Also, Principal Component Analysis (PCA) is employed in LMM to learn the statistically dominant displacements for all video frames. Lei\textit{ et al.}~\cite{mm_lei2020} first used the learning-based method for video motion magnification to extract shape representations from the intermediate layer of neural networks.

\section{Feature Extraction} \label{sec: Feature Representation}
Feature representation plays an important role for MER. With proper extraction, the raw input data of a ME video clip can be represented in a simple and concise form. The two main challenges for the feature descriptor of ME are 1) It should be able to capture the difference in both spatial and temporal domains and 2) It should be able to capture the micro-level differences. 
In our survey, we divide the commonly used feature representations into two main categories: handcrafted feature and learning-based feature.

\subsection{Handcrafted Feature}
Handcrafted feature for face data can be further divided into appearance-based feature and geometry-based feature~\cite{zeng2008survey}. Appearance-based feature represents the intensity
or the texture information of face region data while geometry-based feature describes the face geometrics such as the location of each facial landmark. In the literature, appearance-based feature has shown its effectiveness on dealing with illumination difference and unaligned images~\cite{lin2012human,hu2014real,shen2015gestalt,rivera2012local}. An illustration of handcrafted features is shown in Fig~\ref{fig:handcrafted}. 

\textbf{Local Binary Pattern (LBP)-based Feature}.
LBP-based feature is commonly used in the literature due to the computational simplicity. The LBP was first proposed in~\cite{ojala1994lbp} as a texture operator via thresholding the eight neighbors of each pixel and representing the result with a binary code. An extended version of LBP is described in ~\cite{ojala2002multiresolution} to meet the need of rotation invariance. 

However, for an ME video clip, the dynamic information between different frames is crucial. As a time-domain extension of LBP, LBP-TOP was proposed to deal with dynamic feature analysis~\cite{zhao2007dynamic}. 
Usually, LBP-TOP features are applied with region-based algorithms to improve the robustness of misalignment. Currently, LBP-TOP is reported by most of the existing ME datasets as the baseline evaluation. Due to its computational simplicity, LBP-TOP features are utilized in a variety of different MER frameworks including cross-domain MER and ME spotting~\cite{zong2018domain,zhang2020new,tran2019dense}.




\begin{table*}[!t]
\renewcommand{\arraystretch}{1.3}
\caption{Performance comparison among \textit{macro-to-micro adaptation} methods}
\label{tab: From Macro to Micro}
\centering
\begin{tabular}{|c||c||c||c||c||c||c||c||c| }
\hline
\multirow{2}{*}{Methods} &\multirow{2}{*}{Year}&\multirow{2}{*}{Macro Datasets} &\multirow{2}{*}{Features}&\multirow{2}{*}{Protocol}&\multirow{2}{*}{Micro Datasets} & \multicolumn{3}{c|}{Experimental Results}\\
\cline{7-9}
& & & & & & ACC & F1-score & UAR \\
\hline\hline
\multirow{3}{*}{\cite{mm_xia2020macro}}&
\multirow{3}{*}{2020}&
{CK+}&\multirow{3}{*}{ResNet18}&\multirow{3}{*}{LOSO (cross-dataset)}& CASME II & 0.756 & 0.701 & 0.872  \\
 & &MMI && & SAMM	& 0.741	&0.736	&0.819\\
 &  &Oulu-CASIA & & &SMIC & 0.768 &0.744	&0.861\\
\hline
\multirow{2}{*}{\cite{zhi2019combining}}&
\multirow{2}{*}{2019}&
\multirow{2}{*}{Oulu-CASIA}&\multirow{2}{*}{3D CNN} &\multirow{2}{*}{5-fold cross validation}& CASME II & 0.976 & - & -  \\
 & & & & & SAMM	& 0.974 &-	&-	\\
\hline

\multirow{3}{*}{\cite{zhou2019cross}}&
\multirow{3}{*}{2019}&
\multirow{3}{*}{BU-3DFE }&\multirow{3}{*}{ResNet}&\multirow{3}{*}{LOSO (cross-dataset)}& CASME II & - & 0.761 & 0.755  \\
 & & && & SAMM	& -	&0.448	&0.487\\
 &  & & & &SMIC & -&0.551	&0.546\\
\hline
\cite{jia2018macro} &2018& CK+ &LBP-TOP&-&CASME II &0.836&0.856&0.863\\ 
\hline
\multirow{3}{*}{\cite{wang2020micro}}&
\multirow{3}{*}{2018}&CK+&\multirow{3}{*}{CNN}&\multirow{3}{*}{LOSO}& CASME II & 0.659 & 0.539 & 0.584  \\
 & &Oulu-CASIA & & & SAMM	& 0.485	&0.402	&0.559\\
 &  &Jaffe, MUGFE & & & SMIC & 0.494 &0.496	&-\\
\hline
\multirow{2}{*}{\cite{peng2018macro}}&
\multirow{2}{*}{2018}&
CK+, Oulu-CASIA & \multirow{2}{*}{ResNet10}&\multirow{2}{*}{LOSO (cross-dataset)} &CASME II & 0.757 & 0.650 & -  \\
 & & Jaffe, MUGFE&& & SAMM	& 0.706	&0.540	&-\\
\hline
\cite{ben2018learning} &2018& CK+&DCP-TOP $\&$ HWP-TOP&- &CASME II &0.607	&-	&-\\ 
\hline

\end{tabular}
\end{table*}

\textbf{LBP-TOP variants.} Aside from the use of original LBP-TOP operator, several variants were proposed to meet different needs for MER~\cite{huang2017discriminative,mao2019classroom,guo2019elbptop,hu2018multi,hu2019gender,yu2019lcbp,arango2020mean}. 
Huang \textit{et al.}~\cite{huang2017discriminative} combined the idea of integral projection and texture descriptor like LBP-TOP to bone texture characterization and face recognition. 
Guo \textit{et al.}~\cite{guo2019elbptop}  extended the LBP-TOP operator into two novel binary descriptors: Angular Difference LBP-TOP (ADLBPTOP) and Radial Difference LBP-TOP (RDLBPTOP), respectively. 
Hu \textit{et al.}~\cite{hu2018multi} combined LBP-TOP and learning based features to form a feature fusion for multi-task learning. 
Hu \textit{et al.}~\cite{hu2019gender} also took advantage of Gabor filter and proposed Centralized Gabor Binary Pattern from Three Orthogonal Panels (CGBP-TOP). 
Yu \textit{et al.}~\cite{yu2019lcbp} proposed a new Local Cubes Binary Patterns (LCBP) especially for ME spotting. Instead of using three different plane combinations to represent the spatio-temporal feature~\cite{wu2016unfolding,wu2016time}, they utilized a cube mask to encode the information of eight directional angles for both the space and time domains. 


\textbf{Gradient-Based Feature}.
One of the key challenges for the feature descriptor is to describe the subtle changes in ME sequences. Aside from LBP, local patterns based on gradient have been used for the property that high order gradient would represent the detailed structure information of an image. 
Dalal and Triggs proposed the histogram of gradients (HOG) in 2005~\cite{dalal2005histograms}, which is one of the most commonly used feature descriptors when it comes to object recognition for the ability to specialize the edges in an image. 
The HOG descriptor has the property of geometric invariance and optical invariance. With HOG descriptors, the expression contour feature can be well captured. 

Histogram of Image Gradient Orientation (HIGO) is a variant of HOG. It ignores the magnitude weighting in the original HOG and thus can suppress the illumination effect. Zhang \textit{et al.}~\cite{zhang2020new} utilized both HOG-TOP and HIGO-TOP as the feature descriptors in their work of MER. Tran \textit{et al.}~\cite{tran2019dense} built a spotting network based on LSTM using HOG-TOP and HIGO-TOP features as the input of the network. To better capture the structural changes in ME videos using gradient information, Niu \textit{et al.}~\cite{niu2018LTOGP} proposed a local pattern of Local Two-Order Gradient Pattern (LTOGP). 


\textbf{Optical Flow Based Feature}.
The feature descriptors based on optical flow infer the relative motion information between different frames in order to capture subtle muscle movements for MER.
The idea of optical flow was first introduced by Horn \textit{et al.}~\cite{horn1981determining} to describe the movement of brightness patterns in an image. By utilizing the pixel-wise difference between consecutive frames in a video clip, the motion information of the object in the video can thus be obtained. The basic concept is to find the distance of an identical object in different frames. 

Owing to the fact that optical flow could capture temporal patterns between consecutive frames, one of the most employed architecture is to combine optical flow feature with CNN to further recognize spatial patterns~\cite{xia2020revealing,liong2019shallow,gan2019offapexnet,liu2019neural,zhao2019convolutional,gan2018bi}.
Verburg \textit{et al.}~\cite{verburg2019micro} utilized Histogram of Oriented Optical Flow (HOOF) to encode the subtle changes in the time domain for selected face regions. 
Li \textit{et al.}~\cite{li2019facial,li2018fusing} revisited the HOOF feature descriptor and proposed an enhanced version to reduce the redundant dimensions in HOOF. 
Liong \textit{et al.}~\cite{liong2018biwoof} proposed another feature descriptor based on optical flow, Bi-Weighted Oriented Optical Flow (Bi-WOOF). Bi-WOOF represents a sequence of subtle expressions using only two frames. 
In contrast to HOOF, both the magnitude and optical strain values are used as weighting schemes to highlight the importance of each optical flow so the noisy optical flows with small intensities are reduced. 


However, using histogram of optical flow as feature descriptors may have some flaws. 
When the histogram is used as a feature vector for a classifier, even a slightly shifted version of a histogram will create a huge difference as most classification algorithms use euclidean distance to measure the difference of two images. Happy \textit{et al.}~\cite{happy2017FHOFO} proposed Fuzzy Histogram of Optical Flow Orientations (FHOFO) to collect the motion directions into angular bins based on the fuzzy membership function. It ignores the subtle motion magnitudes during the MEs and only takes the motion direction into consideration. 

\textbf{Other Handcrafted Feature}.
Apart from the abovementioned features, there are other descriptors to capture the distinctive properties embedded in ME videos. 

Li \textit{et al.}~\cite{li2018ltp} defined the local and temporal patterns (LTP) of facial movement. LTP could be extracted from a projection in the PCA space. 
Instead of using LBP-TOP, Pawar \textit{et al.}~\cite{pawar2019micro,sanchez2016comparative} introduced a computationally efficient 3D version of Harris corner function, called Spatio Temporal Texture Map (STTM). 
Lin \textit{et al.} ~\cite{lin2018gabor} proposed a method using spatio-temporal Gabor filters. 



While some approaches divided a face into regions spatially to better capture the low-intensity expression, it is hard to choose an ideal size of division because the division grids directly affect the discriminative feature attained. Zong \textit{et al.}~\cite{zong2018hierarchical} proposed a hierarchical spatial division scheme for spatio-temporal feature descriptor to address this issue.


To leverage the advantage of multiple feature types, there are a few approaches that take more than one type of feature descriptors. Wang \textit{et al.}~\cite{wang2019weighted} first acquired facial LBP-TOP as the base feature, and then calculated the optical-flow as different feature weights.
Zhao \textit{et al.}~\cite{zhao2019improved}, on the other hand, combined LBP-TOP and necessary morphological patches (NMPs). 


\begin{table*}[!t]
\renewcommand{\arraystretch}{1.3}
\caption{Performance comparison among \textit{recognition based on key apex frames} methods}
\label{tab: only using key frames}
\centering

\begin{tabular}{|c||c||c||c||c||c||c||c||c|c| }
\hline
\multirow{2}{*}{Methods} &\multirow{2}{*}{Year}&\multirow{2}{*}{Key Frames} &\multirow{2}{*}{Features}&\multirow{2}{*}{Protocol}&\multirow{2}{*}{Datasets} & \multicolumn{3}{c|}{Experimental Results}\\
\cline{7-9}
& & & & & & ACC & F1-score & UAR  \\
\hline\hline
\cite{zhong2020facial} &2020& Key Frame (SSIM) &dual-cross patterns (DCP)&-&CASME II &0.687	&-	&-	\\ 
\hline
\multirow{3}{*}{\cite{xia2020revealing}}&
\multirow{3}{*}{2020}&
\multirow{3}{*}{Onset $\&$ Apex}&\multirow{3}{*}{RCN}&\multirow{3}{*}{LOSO}& CASME II & - & 0.856 & 0.812  \\
 & & & & &SAMM	& -	&0.765	&0.677	\\
 &  & & & &SMIC &- &0.658	&0.660	\\
\hline
\multirow{3}{*}{\cite{gan2019off}}&
\multirow{3}{*}{2019}&
\multirow{3}{*}{OFF $\&$ Apex}&\multirow{3}{*}{optical flow $\&$ CNN}&\multirow{3}{*}{LOSO}& CASME II* & 0.883 & 0.870 & -   \\
 & & && & SAMM*	& 0.682	&0.542	&-	\\
 &  & && & SMIC* &0.677 &0.671	&-	\\
\hline
\multirow{3}{*}{\cite{zhao2019convolutional}}&
\multirow{3}{*}{2019}&
\multirow{3}{*}{Onset $\&$ Apex}&\multirow{3}{*}{optical flow}&\multirow{3}{*}{-}& CASME II & 0.870 & - & -   \\
 & & && & SAMM	& 0.698	&-	&-	\\
 &  & && & SMIC &0.702 &-	&-	\\
\hline
\multirow{2}{*}{\cite{liong2018less}}&
\multirow{2}{*}{2018}&
\multirow{2}{*}{Apex }&\multirow{2}{*}{Bi-WOOF}&\multirow{2}{*}{-}& CASME II & 0.589 & 0.610 & -  \\
 & & && & SAMM	& 0.622	&0.620	&-	\\
\hline
\multirow{3}{*}{\cite{zhou2019cross}}&
\multirow{3}{*}{2019}&
\multirow{3}{*}{Apex }&\multirow{3}{*}{ResNet}&\multirow{3}{*}{LOSO (cross-dataset)} & CASME II & - & 0.761 & 0.755  \\
 & & & & &SAMM	& -	&0.448	&0.487	\\
 &  & && & SMIC & -&0.551	&0.546\\
\hline
\multirow{2}{*}{\cite{peng2018macro}}&
\multirow{2}{*}{2018}&
\multirow{2}{*}{Apex }&\multirow{2}{*}{ResNet10}&\multirow{2}{*}{LOSO (cross-dataset)}& CASME II & 0.757 & 0.650 & -   \\
 & & && & SAMM	& 0.706	&0.540	&-	\\
\hline
\end{tabular}
\begin{tablenotes} 
 \footnotesize
\item * on upper-right of the name of datasets means the labels of positive, negative and surprise were used in the experiments.
\end{tablenotes} 
\end{table*}

\subsection{Learning based Feature}
In recent years, deep learning has shown a substantial improvement in computer vision tasks~\cite{deng2009imagenet,ren2015faster,wang2017background,hua2015computer,abousaleh2016novel,hsieh2019fashionon,susanto2020emotion,cheng2018ai+}. Convolutional Neural Networks (CNNs, or ConvNet) have been considered as the most widely used ways for visual feature extraction with discriminative ability~\cite{deng2009imagenet}.


Since MEs are sequences of images, 2D CNN and 3D CNN are useful for their representative feature learning. 2D CNN is generally used on image data, which was first introduced in LeNet5~\cite{LeNet5}. It is called two dimensional CNN because the kernel slides along two dimensions on the data. Input and output data in 3D CNN are four-dimensional since the kernel moves in three directions. 3D CNN is mostly used on 3D image data, e.g. MRI, CT scans, and videos~\cite{ji20123d,chen20193d,hua2015computer_20}. 

\textbf{Shallow learning.} Deep learning-based methods require large-scale data to train the model. However, current public ME datasets are mostly limited and imbalanced, and tend to cause over-fitting issues when directly applying ConvNet. Most works usually adopt shallow and lightweight layers for MER. Dual Temporal Scale Convolutional Neural Network (DTSCNN)~\cite{peng2017dual} was the first end-to-end middle-size neural network for MER. The DTSCNN designs two temporal channels, where each channel only has 4 convolutional layers and 4 pooling layers to partially avoid over-fitting. Peng \textit{et al.}~\cite{peng2018macro} adopted a simplified version of ResNet, ResNet10 for representative learning. Takalkar \textit{et al.}~\cite{takalkar2017image} employed five convolutional layers, three max-pooling layers, and three fully-connected layers. 
Zhao \textit{et al.}~\cite{zhao2019convolutional} used four convolutional layers and three pooling layers for capturing discriminative and high-level ME features. The 1$\times$1 convolutional layer is added right after the input layer to increase the non-linear expression of input data without increasing the computational load of the model. Micro-attention~\cite{wang2020micro} was built with 10 residual blocks, whose micro-attention unit is an extension of ResNet. The short-cut connection designed for identity mapping can reduce the degradation problem. By learning the spatial attention of feature maps, the network can focus on the facial subtle movements. 


\begin{figure}[!t]
\centering
\includegraphics[scale = 0.1]{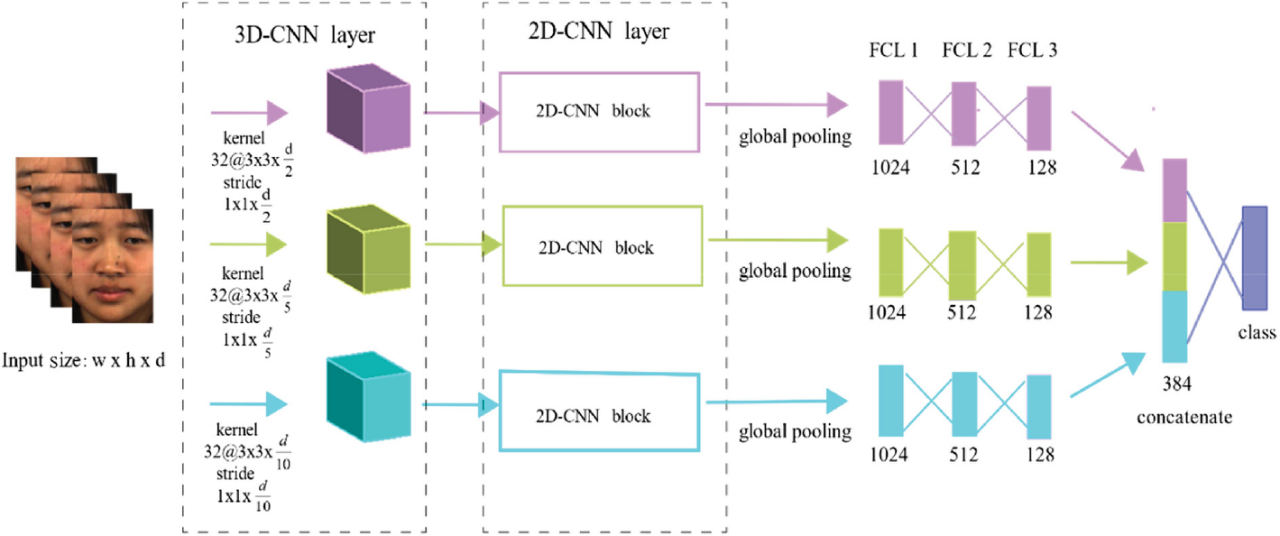}
\caption{The combination structure of the 2D CNN and 3D CNN in TSNN‐LF for MER~\cite{wu2020tsnn}.}
\label{fig:2D CNN and 3DCNN}
\end{figure}

\textbf{Two-step learning.} There are many works that first extract spatial features among all frames, then use recurrent convolutional layer or LSTM module to explore their temporal correlation. Due to the small amount of training samples, many learning based works~\cite{xia2020revealing,khor2018enriched,verburg2019micro} include handcrafted features (e.g., optical flow, HOOF) to give a higher signal-to-noise (SNR) ratio in comparison to using raw pixel data.

The optical flow map and its related extension algorithms, e.g., HOOF, are commonly used for extracting motion deformations of facial regions and have shown good performance~\cite{verburg2019micro,xia2020revealing,khor2018enriched,verburg2019micro}. RCN-F~\cite{xia2020revealing} first extracted an optical flow map from the located onset and apex frames followed by a recurrent convolutional network (RCN). The RCN is composed of three parameter-free modules (i.e., wide expansion, shortcut connection, and attention unit) to enhance the representational ability. Enriched Long-term Recurrent Convolutional Network (ELRCN)~\cite{khor2018enriched} also used the optical flow as input data for two learning modules: dimension enrichment and temporal dimension enrichment. 


\textbf{3D CNN.} In 2D CNN, convolution and pooling operations are applied on 2D feature maps which is lacking of motion information. To better preserve temporal information of the input signals effectively, 3D convolution has been proved its ability for capturing more representative features in the spatio-temporal aspect~\cite{ji20123d}. Considering that MEs occur only in a short period among dynamic facial movements and can be explicitly analyzed by detecting their constituent temporal segments, several MER works~\cite{li2019micro,zhi2019combining,reddy2019spontaneous,wu2020tsnn} used 3D CNN to extract the facial temporal movements. 

Li \textit{et al.}~\cite{li2019micro} applied the 12-layer 3D flow-based CNNs model for MER, which extracts motion flow information arising from subtle facial movements. To better represent the dynamic and appearance features of MEs, optical flow and gray scale frames are combined as input data. The 3D convolution adopts 3$\times$3$\times$3 kernels to represent the spatial changes at each local region. Zhi \textit{et al.}~\cite{zhi2019combining} employed 3D CNN for self-learning feature extraction. All convolutional kernels are set to the size of 3$\times$5$\times$5, where 3 is the temporal depth and 5$\times$5 is the size of the spatial receptive field by taking consideration of efficiency and computational complexity. Reddy \textit{et al.}~\cite{reddy2019spontaneous} proposed two 3D CNN models for spontaneous MER, i.e., MicroExpSTCNN and MicroExpFuseNet, by exploiting the spatio-temporal information in the CNN framework. 

Nevertheless, applying deep 3D CNN from scratch significantly increases the computational cost and memory demand. Besides, most popular 3D networks, e.g., C3D and P3D ResNet, are trained on Sports 1-M, that is an action database and is very different from MEs. Hence, it still needs to investigate the trained deep neural networks for ME feature extraction, which would boost the performance of the MER task~\cite{zong2018domain}. TSNN~\cite{wu2020tsnn} designed a three‐stream combining 2D and 3D CNN to extract expression sequence feature. The advantages of single 3D kernel sizes and multiple 3D kernel combination have been made full use in the proposed framework to improve the MER performance. The architecture of TSNN is shown in Fig.~\ref{fig:2D CNN and 3DCNN}.

\section{Micro-expression Spotting}
\label{sec:ME spotting}
Micro-expression spotting refers to locate the segments of micro movements for a given video. There are less publications compared to MER in the literature~\cite{gupta2018exploring} due to the difficulty of discovering subtle difference in a short duration (see Fig.~\ref{fig:publications}). However, ME spotting is a vital step for automatic ME analysis and has attracted more and more attention recently, since proper spotting can decrease the redundant information for further recognition. Table~\ref{tab: spotting} compares the \textit{state-of-the-art} micro-expression spotting methods.
Several earlier studies on automatic facial ME analysis primarily focused on distinguishing facial micro-expressions from macro-expressions~\cite{tran2020micro,shreve2009towards,shreve2011macro}. Shreve \textit{et al.}~\cite{shreve2009towards,shreve2011macro} used an optical flow method for automatic ME spotting on their own database. However, their database contains 100 clips of posed MEs, which were obtained by asking participants to mimic some example videos that contain micro-expressions. 

Tran \textit{et al.}~\cite{verburg2019micro} first introduced deep sequence model for ME spotting. LSTM shows its effectiveness on using both local and global correlations of the extracted features to predict the score of the ME apex frame.
Li \textit{et al.}~\cite{li2017towards} first proposed an ME spotting method which was effective on spontaneous ME datasets. A spotting ME convolutional network~\cite{zhang2018smeconvnet} was designed for extracting features from video clips, which is the first time that deep learning is used in ME spotting.

The co-occurrence of macro- and micro-expressions are common in real life. An automatic spotting system for micro- and macro-expressions was designed by~\cite{zhang2020spatio}. 
Based on the fact that micro- and macro-expressions have different intensities, a multi-scale filter is used to improve the performance.

\textbf{Onset and offset detection.} While it is relatively easier to identify the peaks and valleys of facial movements, the onset and offset frames are much more difficult to determine. Locating the onset and offset frames is crucial for real-life situations where facial movements are continuously changing. CASME II and SAMM provide the onset and offset frame ground-truth, which can be helpful for model training. 

\textbf{Apex frame spotting.} Liong \textit{et al.}~\cite{liong2015automaticapex} introduced an automatic apex frame spotting method and this strategy was also adopted in~\cite{zhou2019cross,liong2018less,gan2018bi}. The LBP feature descriptor was first employed to encode the features of each frame, then a \textit{divide-and-conquer} methodology was exploited to detect the frames with peak facial changes as apex frame. 

Most of the aforementioned ME spotting methods were conducted on public ME lab-controlled datasets. However, ME spotting in real-world scenes with different environmental factors is still an open issue. More ME datasets containing long video and real-world scenes are essential for further research.

\section{MER Methodology}
\label{sec:ME approches}
\subsection{Macro-to-Micro Adaptation}

As mentioned in the previous section, how to solve automatic labeling and recognition problems for MEs is a challenging task under the condition of small number of labeled training ME samples available. Meanwhile, there are large amounts of macro-expression databases~\cite{FERSurvey2019,weber2018survey}, each of which consists of vast labeled training samples compared with micro-expression databases. Thus, how to take advantage of the macro-expression databases for MER has become an important direction for research. Table~\ref{tab: From Macro to Micro} summarizes the \textit{macro-to-micro adaptation} methods.

\textbf{Transfer-learning based method} has proved to be efficient in applying deep CNN on small databases~\cite{tan2018survey}. Thus, by using the idea of transfer learning, it is reasonable to take advantage of the quantitative superiority of macro-expression to recognize the micro-expression~\cite{peng2018macro, jia2018macro,wang2020micro,zhi2019combining, zhou2019cross}. 
In~\cite{jia2018macro}, the macro-expression features and micro-expression features are considered gallery and probe features respectively and constructed as training matrix, where the gallery features can be transformed to the probe features. The two different features were then projected into a joint subspace, where they are associated with each other. The nearest neighbor (NN) classifier was used to classify the probe micro-expression samples at the last step.
Zhi \textit{et al.}~\cite{wang2020micro} pretrained the 3D CNN on macro-expression database Oulu-CASIA~\cite{zhao2011facial}, and then the pre-trained model was transferred to the target micro-expression domain. The experimental results show 3.4\% and 1.6\% in MER performance higher than the model without transfer learning, respectively.

Xia \textit{et al.}~\cite{mm_xia2020macro} imposed a loss inequality regularization to make the output of MicroNet converge to that of MacroNet. In~\cite{ben2018learning}, macro-expression images and micro-expression sequences were encoded by proposed hot wheel patterns (HWP), Dual-cross patterns (DCP-TOP) and HWP-TOP, respectively. The coupled metric learning algorithm was employed to model the shared features between micro-expression samples and macro-information.
In~\cite{wang2020micro}, the original residual network (without micro-attention units) was first initialized with the ImageNet database. Then, to narrow the gap between object recognition (ImageNet) and facial expression recognition, the network was further pre-trained on several popular macro-expression databases, including CK+~\cite{lucey2010extended}, Oulu-CASIA NIR $\&$ VIS~\cite{zhao2011facial}, Jaffe~\cite{lyons1998coding}, and MUGFE~\cite{aifanti2010mug}. Finally, the residual network together with micro-attention units is fine-tuned with micro-expression databases, including CASME II, SAMM and SMIC. Similarly, Jia \textit{et al.}~\cite{peng2018macro} proposed a macro-to-micro transformation network. ResNet10 pre-trained on ImageNet dataset was fine-tuned on macro-expression datasets first and then on the micro-expression datasets (CASME II, SAMM).

\textbf{Knowledge distillation} strategy is also used for MER. Sun \textit{et al.}~\cite{sun2020dynamic} proposed a multi-task teacher network containing AUs recognition and facial view classification on FERA2017 dataset~\cite{valstar2017fera}, then the learned knowledge was distillated to a shallow student network for MER. In SA-AT~\cite{zhou2019cross}, the larger scale of macro-expression data were served as auxiliary database to train a teacher model, then the teacher model was transferred to train the student model on micro-expression databases with limited samples. 
To narrow the gap between macro- and micro-expressions, a style aggregated strategy was used to transform micro-expression samples from different macro-expression datasets to generate an aggregated style via CycleGAN~\cite{zhu2017unpaired}.


\begin{figure}[!t]
\centering
\includegraphics[scale = 0.2]{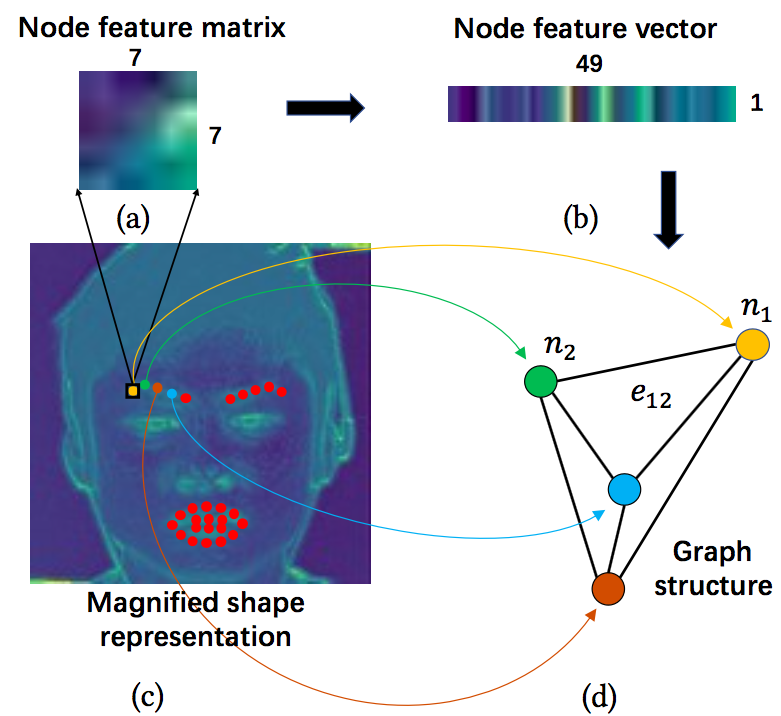}
\caption{Sample facial graph built in Graph-TCN~\cite{mm_lei2020}.}
\label{fig:facial graph building}
\end{figure}

\subsection{Recognition based on Key Apex Frames}

Apex frame, which is the instant indicating the most expressive emotional state in a video sequence, is effective to classify the emotion in that particular frame. The apex frame portrays the highest intensity of facial motion among all frames. The apex occurs when the change in facial muscle reaches the peak or the highest intensity of the facial motion. 
Many works~\cite{gan2019off,xia2020revealing} extracted features of the apex frame as feature descriptor. 

Existing ME datasets, e.g., CASME II and SAMM, provide the index annotation of apex frames, while SMIC-HS database does not release the apex information. Thus, automatic apex frame spotting~\cite{liong2015automaticapex,zhou2019cross,liong2018less,gan2018bi} is necessary to be applied to each video to acquire the location of the apex frame. Table \ref{tab: only using key frames} summarizes the \textit{state-of-the-art} key apex frames based MER approaches.

\textbf{Using single apex frame}. Considering the subtle motion change of ME frames in video, several works emphasized the use of a single apex frame to minimize the redundancy of the repetitive input frames. Using apex frame to represent the whole ME sequence was expected to reduce the computational complexity for feature learning~\cite{peng2018macro,gan2018bi,liong2018less,zhou2019cross}. The experimental results of \cite{sun2020dynamic} showed that the effect of only using apex frame is better than onset-apex-offset sequence and the whole video.

\begin{table*}[!htp]
\centering
\begin{threeparttable} 
\renewcommand{\arraystretch}{1.3}
\caption{Performance comparison on CASME II dataset}
\label{tab:handcrafted features on CASME II} 
\centering
\begin{tabular}{|c||c||c||c||c||c|c|c|}
\hline
\multirow{2}{*}{Methods} &\multirow{2}{*}{Year}&\multirow{2}{*}{Features} &\multirow{2}{*}{Classifier}&\multirow{2}{*}{Protocol} & \multicolumn{3}{c|}{Experimental Results}\\
\cline{6-8}
& & & & & ACC & F1-score & UAR \\
\hline
~\cite{guo2019elbptop}
& 2019
& ELBPTOP
& SVM
& LOSO
& 0.739 & 0.690 & -\\
\hline
~\cite{guo2019elbptop}$^*$
& 2019
& ELBPTOP
& SVM
& LOSO
& 0.796 & 0.660 & -\\
\hline
~\cite{hu2019gender}
& 2019
& CGBP-TOP
& Gaussian kernel function
& LOSO
& 0.658 & - & -\\

\hline
~\cite{li2019facial}
& 2019
& revised HOOF + CNN
& MLP
& LOSO
& 0.580 & - & - \\
\hline
~\cite{Allaert2019lmp}
& 2019
& LMP
& SVM
& LOSO
& 0.702 & - & - \\
\hline
~\cite{pawar2019micro}
& 2019
& spatiotemporal texture map
& SVM
& k-fold cross validation
& 0.800& 0.890 & -\\
\hline
~\cite{asmara2019drmf}$^{*}$
& 2019
& landmark point feature
& MLP
& LOSO
& 0.793 & - & - \\
\hline
~\cite{huang2017discriminative}
& 2018
& discriminative STLBP
& SVM + Gentle adaboost
& LOSO
& 0.648 & - & -\\
\hline
~\cite{lu2018motion}
& 2018
& FMBH
& SVM
& LOSO
& 0.643 & - & - \\
\hline
~\cite{zong2018hierarchical}
& 2018
& hierarchical ST descriptor
& GSL model
& LOSO
& 0.639 & 0.613 & - \\
\hline
\hline
~\cite{xia2020revealing} $^*$
& 2020
& optical flow + RCN
& MLP
& LOSO
& - & 0.856 & 0.813\\
\hline
~\cite{liong2019shallow}$^*$
& 2019
& optical flow+STSTNet
& MLP
& LOSO
& - & 0.680 & 0.701\\
\hline
~\cite{gan2019offapexnet}$^*$
& 2019
& optical flow + CNN
& MLP
& LOSO
& 0.883 & 0.897 & -\\
\hline
~\cite{liu2019neural}$^*$
& 2019
& optival flow + CNN
& MLP
& LOSO
&  - & 0.829 & 0.820\\
\hline
~\cite{Zhou2019dual}$^*$
& 2019
& optical flow + CNN
& MLP
& LOSO
& - & 0.862 & 0.856\\
\hline
~\cite{wang2019weighted}
& 2019
& LBP-TOP + optical flow
& SVM
& LOSO
& 0.696 & - & - \\
\hline
~\cite{zhao2019convolutional}
& 2019
& optical flow + CNN
& MLP
& LOSO
& 0.870 & - & - \\
\hline
~\cite{verma2019learnet}
& 2019
& dynamic imaging + CNN
& MLP
& k-fold cross validation
& 0.766 & - & - \\
\hline
~\cite{zhong2020facial}
& 2019
& local ROI+DCP
& chi-square distance
& -
& 0.687& - & -\\
\hline
~\cite{hu2018multi}
& 2018
& LGBP-TOP + CNN
& MLP
& LOSO
& 0.662 & - & - \\
\hline
~\cite{lin2018gabor}
& 2018
& frame difference + ST Gabor filter
& SVM
& LOSO
& 0.553 & - & - \\
\hline
\hline
\cite{xie2020assisted}&
2020&
3D CNN&
Softmax &
LOSO $\&$ LOVO&
0.561 & 0.394 & -\\
\hline
\cite{mm_lei2020}&
2020&
Graph-TCN&
Softmax &
LOSO& 
 0.740 & 0.725 & -  \\
 \hline
\cite{mm_xia2020macro}&
2020&
ResNet18&
Triplet loss&
LOSO& 
0.756 & 0.701 & 0.872 \\
\hline
\cite{li2019micro}&
2019&
3D flow-based CNN&
SVM&
LOSO&
0.591& - & -\\
\hline
\cite{zhi2019combining}&
2019&
3D CNN&
SVM &
5-fold cross validation&
0.976 & - & -  \\
\hline
\cite{peng2018macro}&2018&ResNet10&-&LOSO& 0.757 & 0.650 & -   \\
\hline 
\cite{wang2020micro}&2018&CNN&Softmax&LOSO & 0.659 & 0.539 & 0.584\\
\hline

\end{tabular}

\begin{tablenotes} 
\centering
\item The * on upper-right of the methods mean only three labels of positive, negative and surprise were used in the experiments
.\end{tablenotes} 

\end{threeparttable} 
\end{table*}

\textbf{The optical flows from the onset and apex frames} are commonly used for extracting motion deformations of facial regions and can achieve good performance in subject independent evaluation~\cite{xia2019spatiotemporal,liong2019shallow,zhao2019convolutional,xia2019cross}. Zhou \textit{et al.}~\cite{Zhou2019dual} took the mid-position frames in ME samples based on the observation that most apex frame existing in the middle segment of a data sequence. OFF-ApexNet~\cite{gan2019off} adopted onset and apex frames to represent the ME details and extracted optical flow to encode the motion flow features based on two chosen frames. The two streams are combined in a fully-connected layer for MER. The similar strategy has been used in~\cite{xia2020revealing}. The obtained onset and apex frames are combined with the corresponding optical flow map as input of the proposed MER model. 
Zhao \textit{et al.}~\cite{zhao2019convolutional} used the Total Variation-L1 (TV-L1) optical flow algorithm to calculate the motion information between the onset frame and the apex frame. Then, the corresponding optical flow feature map is obtained and fed into subsequent deep network learning. 

\textbf{Others.} Key frames of each local region are computed by SSIM in~\cite{zhong2020facial}, and then the dual-cross patterns (DCP) are applied to get the final feature vectors.


\begin{table*}[!htp]
\centering
\begin{threeparttable} 
\renewcommand{\arraystretch}{1.3}
\caption{Performance comparison on SMIC dataset}
\label{tab:handcrafted features on smic} 
\centering
\begin{tabular}{|c||c||c||c||c||c|c|c|}
\hline
\multirow{2}{*}{Methods} &\multirow{2}{*}{Year}&\multirow{2}{*}{Features} &\multirow{2}{*}{Classifier}&\multirow{2}{*}{Protocol} & \multicolumn{3}{c|}{Experimental Results}\\
\cline{6-8}
& & & & & ACC & F1-score & UAR \\
\hline
~\cite{guo2019elbptop}
& 2019
& ELBPTOP
& SVM
& LOSO
& 0.691 & 0.650 & 0.660\\
\hline
~\cite{hu2019gender}
& 2019 
& CGBP-TOP
& Gaussian kernel function
& LOSO
& 0.594 & - & - \\
\hline
~\cite{Allaert2019lmp}
& 2019
& LMP
& SVM
& LOSO
& 0.674 & - & - \\
\hline
~\cite{huang2017discriminative}
& 2018
& discriminative STLBP
& SVM + Gentle adaboost
& LOSO
& 0.634 & -	& - \\
\hline
~\cite{lu2018motion}
& 2018
& FMBH
& SVM
& LOSO
& 0.683 & - & - \\
\hline
~\cite{liong2018biwoof}
& 2018
& Bi-WOOF
& SVM
& LOSO
& 0.620 & - & - \\
\hline
~\cite{happy2017FHOFO}
& 2018
& FHOFO
& SVM
& LOSO
& 0.512 & 0.518 & - \\
\hline
~\cite{zong2018hierarchical}
& 2018
& hierarchical ST descriptor
& GSL model
& LOSO
& 0.604 & 0.613 & - \\

\hline
\hline
~\cite{xia2020revealing}
& 2020
& optical flow + RCN
& MLP
& LOSO
& - & 0.658 & 0.660\\
\hline
~\cite{wang2019weighted}
& 2019
& LBP-TOP + optical flow
& SVM
& LOSO
& 0.717 & - & - \\
\hline
~\cite{liong2019shallow}
& 2019
& optical flow + STSTNet
& MLP
& LOSO
& - & 0.659 & 0.681\\
\hline
~\cite{liu2019neural}
& 2019
& optical flow + CNN
& MLP 
& LOSO
& - & 0.746 & 0.753\\
\hline
~\cite{Zhou2019dual}
& 2019
& optical flow + CNN
& MLP
& LOSO
& - & 0.664 & 0.673 \\
\hline
~\cite{gan2019offapexnet}
& 2019
& optical flow + CNN
& MLP
& LOSO
& 0.677 & 0.671 & - \\
\hline
~\cite{zhao2019convolutional}
& 2019
& optical flow + CNN
& MLP
& LOSO
& 0.698 & - & - \\
\hline
~\cite{verma2019learnet}
& 2019
& dynamic imaging + CNN
& MLP
& k-fold
& 0.911 & - & - \\
\hline
~\cite{hu2018multi}
& 2018
& LGBP-TOP + CNN
& MLP
& LOSO
& 0.651 & -	& - \\
\hline
~\cite{lin2018gabor}
& 2018
& frame difference + ST Gabor filter
& SVM
& LOSO
& 0.545 & - & - \\
\hline
\hline
\cite{mm_xia2020macro}&
2020&
ResNet18&
Triplet loss&
LOSO& 
0.768 & 0.744 & 0.861 \\
\hline
\cite{li2019micro}&
2019&
3D flow-based CNN&
SVM&
LOSO&
0.555& - & -\\
\hline
\cite{reddy2019spontaneous}&
2019&
3DCNN &
Softmax&-&
0.650	&-	&-\\
\hline
\cite{wang2020micro}&2018&CNN&Softmax&LOSO & 0.494 & 0.496 & -\\
\hline
\end{tabular}
\end{threeparttable} 
\end{table*}

\subsection{Recognition based on Facial Action Units}

As physical subtle variations in facial expressions, AUs analysis are crucial because they constitute essential signals to be understood.
Emotion states may be conceptually similar in terms of the difference in facial muscle movements (e.g., fear and disgust). There are many works focusing on facial AU recognition~\cite{luself,wang2019multi}, including AU occurrence detection~\cite{aurcnn,aucrossdomain,aumultilabel,eacnet} and AU intensity estimation~\cite{fan2020AUintensity,li2020learning}. Some surveys compare recent AUs recognition works, e.g.,~\cite{zhi2020_AUSurvey,martinez2017_AUSurvey}.

Since the occurrence of AUs are strongly correlated, AU detection is usually considered as a multi-label learning problem. 
Several works considered the relationships among AUs and modeled AU interrelations to improve recognition accuracy~\cite{liu2020relation,li2019semantic,chen2019multi}. However, most works rely on probabilistic graphical models with manually extracted features~\cite{autraditional, autraditional2}. Given that the graph has the ability of handling multi-relational data~\cite{chen2019multi}, Liu \textit{et al.}~\cite{liu2020relation} proposed the first work that employed GCN to model AU relationship. The cropped AU regions by EAC-Net~\cite{eacnet} are fed into GCN as nodes, after that the propagation of the graph is determined by the relationship of AUs.

AU intensities are annotated by appending five-point ordinal scale (A–E for minimal-maximal intensity)~\cite{martinez2017_AUSurvey}. The intensity levels are not uniform intervals. To infer the co-occurences of AUs, it can be formulated into a heatmap regression-based framework. Fan \textit{et al.}~\cite{fan2020AUintensity} modeled AU co-occurring patterns among feature channels, where semantic descriptions and spatial distributions of AUs are both encoded.

Further insights into locating AUs could possibly provide even better discrimination between types of MEs. Some public ME datasets, e.g., CASME II and SAMM, contain emotion classes as well as AUs annotation, which makes the relationship construction between ME classes and AUs become achievable. Until now, there is few literature exploring automatic MER based on AUs~\cite{wang2015micro,liu2015main,li2019semantic,lo2020mer-gcn}. 
Wang \textit{et al.}~\cite{wang2015micro} defined 16 ROIs based on AUs and features extracted for each ROI can further boost the MER performance. Liu \textit{et al.}~\cite{liu2015main} partitioned the facial region into 36 ROIs from 66 feature points and proposed a ROI-based optical feature, MDMO for MER. However, such predefined rules for modeling relationships among AUs may lead to limited generalization.
Li \textit{et al.}~\cite{li2019semantic} used a structured knowledge-graph for AUs and integrated a Gated GNN (GGNN) to generate enhanced AU representation. This is the first work that integrates AU detection with MER. Lo \textit{et al.}~\cite{lo2020mer-gcn} proposed MER-GCN, which was the first end-to-end AU oriented MER architecture based on GCN, where GCN layers are able to discover the dependency laying between AU nodes for MER. Graph-TCN~\cite{mm_lei2020} utilized the graph structure for node and edge feature extraction, where the facial graph construction is shown in Fig.~\ref{fig:facial graph building}. Sun \textit{et al.}~\cite{sun2020dynamic} proposed a knowledge transfer strategy that distills and transfers AUs information for MER.

\section{Loss Function} \label{sec: Loss Function Design}
\subsection{Ranking Loss}
Ranking Loss, such as \textit{Contrastive Loss}, \textit{Margin Loss},\textit{ Hinge Loss} or \textit{Triplet Loss}, is widely used in MER~\cite{zhang2020new, li2020deep,verma2019learnet,li2019spotting,merghani2020adaptive,takalkar2020manifold,mm_xia2020macro,lim2017fuzzy}. The objective of ranking loss is to predict relative distances between inputs, which is also called metric learning~\cite{pan2019furniture}. Lim and Goh~\cite{lim2017fuzzy} proposed Fuzzy Qualitative Rank Classifier (FQRC) to model the ambiguity in MER task by using a multi-label rank classifier. Xia \textit{et al.}~\cite{mm_xia2020macro} imposed an loss inequality regularization in triplet loss to make the output of MicroNet converge to that of MacroNet.
The hinge loss is used for maximum-margin classification, most notably for Support Vector Machines (SVMs). 
SVM classify training data points by finding a discriminative hyperplane function according to different kernels such as Linear kernel, Polynomial kernel, and Radial Basis Function (RBF) kernel. Each different kernel has its own advantage for different separability and order of the dataset. 
Li \textit{et al.}~\cite{li2019spotting} split each video into 12 feature ensembles, which represent the ME local movements, and SVM was employed to classify these local ROIs. LEARNet~\cite{verma2019learnet} employed RankSVM~\cite{smola2004tutorial} to compute the frame scores in a video. 
To reduce computation time, Sequential Minimal Optimization (SMO), one of the computationally fastest methods of evaluating linear SVMs, was used in~\cite{merghani2020adaptive}.

Overall, SVM is one of the most well-applied classifiers for MER. However, they show poor performances when the feature dimension is far greater than the training sample number.

\begin{table*}[!t]
\centering
\begin{threeparttable} 
\renewcommand{\arraystretch}{1.3}
\caption{Performance comparison on SAMM dataset}
\label{tab:handcrafted features on samm} 
\centering
\begin{tabular}{|c||c||c||c||c||c|c|c|}
\hline
\multirow{2}{*}{Methods} 
&\multirow{2}{*}{Year}
&\multirow{2}{*}{Features} &\multirow{2}{*}{Classifier}
&\multirow{2}{*}{Protocol}
&\multicolumn{3}{c|}{Experimental Results}\\
\cline{6-8}
& & & & & ACC & F1-score & UAR \\
\hline
~\cite{guo2019elbptop}
& 2019
& ELBPTOP
& SVM
& LOSO
& 0.634 & 0.480 & -\\
\hline
~\cite{guo2019elbptop}$^*$
& 2019
& ELBPTOP
& SVM
& LOSO
& - & 0.780 & 0.720\\
\hline
~\cite{asmara2019drmf}$^{*}$
& 2019
& landmark point feature
& MLP
& LOSO
& 0.746 & - & - \\
\hline
\hline
~\cite{xia2020revealing}$^*$
& 2020
& optical flow + RCN
& MLP
& LOSO
& - & 0.765 & 0.677\\
\hline
~\cite{liong2019shallow}$^*$
& 2019
& optical flow + STSTNet
& MLP
& LOSO
& - & 0.838 & 0.869\\
\hline
~\cite{liu2019neural}$^*$
& 2019
& optical flow + CNN
& MLP
& LOSO
& - & 0.775 & 0.715\\
\hline
~\cite{Zhou2019dual}$^*$
& 2019
& optical + CNN
& MLP
& LOSO
& - & 0.587 & 0.566\\
\hline
~\cite{gan2019offapexnet}$^*$
& 2019
& optical + CNN
& MLP
& LOSO
& 0.682 & 0.542 & -\\
\hline
~\cite{zhao2019convolutional}
& 2019
& optical + CNN
& MLP
& LOSO
& 0.702 & - & - \\
\hline
\hline
\cite{mm_xia2020macro}&
2020&
ResNet18&
Triplet loss&
LOSO& 
0.741 & 0.736 & 0.819 \\
\hline
\cite{xie2020assisted}&
2020&
3D CNN&
Softmax &
LOSO $\&$ LOVO&
0.523 & 0.357 & -\\
\hline
\cite{zhi2019combining}&
2019&
3D CNN&
SVM &
5-fold cross validation&
0.974 & - & -  \\
\hline
\cite{wang2020micro}&2018&CNN&Softmax&LOSO & 0.485 & 0.402 & 0.559\\
\hline 
\end{tabular}
\begin{tablenotes} 
\centering
\item The * on upper-right of the methods mean only three labels of positive, negative and surprise were used in the experiments
.\end{tablenotes} 
\end{threeparttable} 
\end{table*}

\subsection{Cross-Entropy Loss}
Unlike traditional methods, where the feature extraction step and the feature classification step are independent, deep networks can perform MER in an end-to-end way. Face recognition system, e.g., Deepface~\cite{taigman2014deepface}, first adopted cross-entropy based softmax loss for facial feature learning. Specifically, softmax loss is the most commonly used function that minimizes the cross-entropy between the estimated class probabilities and the ground-truth distribution. Most works directly applied softmax loss in the MER network~\cite{verburg2019micro,xia2020revealing,gan2019off}. For example, the softmax function was employed followed by an LSTM network which consists of two LSTM layers, each with 12 dimensions~\cite{verburg2019micro}. 

\subsection{Others}
Since the MER task suffers from not only high inter-class similarity but also high intra-class variation, instead of simply employed SVM or softmax loss, several works have proposed novel loss layers for MER~\cite{lalitha2020micro,lin2017focal,lai2020real,xie2019adaptive}. 

Inspired by the center loss, which penalizes the distance between deep features and their corresponding class centers, Lalitha \textit{et al.}~\cite{lalitha2020micro} combined cross-entropy loss and center loss to improve the discriminative energy of the deeply learned features. The focal loss~\cite{lin2017focal} was utilized in MACNN~\cite{lai2020real} to overcome the sample imbalance challenge. Compared with cross-entropy loss, focal loss is more suitable to solve the problem of small sample classification or imbalance between samples. Besides, Xie \textit{et al.}~\cite{xie2019adaptive} proposed a \textit{feature loss} metric to use the complementary information of handcrafted and deep features during training.

\section{Experiments} \label{sec: Overall Comparison/discussion}

\subsection{Evaluation}
\subsubsection{Evaluation Protocols}
The commonly used evaluation protocols for MER are \textit{k}-fold cross-validation, leave-one-subject-out (LOSO) and leave-one-video-out (LOVO). \textit{k}-fold cross-validation repeats random sub-sampling validation. For LOSO validation, the model leaves out all samples of one single subject for the performance evaluation, and all other data are used as training data. The overall performance is then averaged from all folds. Similar to LOSO, LOVO validation protocol requires the model to pick up the frames from one video for validation purpose while all other data are sampled for training. From our review, the LOSO is the most widely used.

There are limitations for the above protocols. With severe class-imbalanced ME data, the effectiveness of \textit{k}-fold is easily influenced. The same problem goes with LOVO, which can introduce additional biases on certain subjects that have more representations during the evaluation process. Moreover, the evaluation results may be over-estimated due to the large training data. For LOSO, the intrinsic ME dynamics of each subject may be limited since the intensity and manner of MEs may differ from person to person. 

\subsubsection{Evaluation Metrics}
The popularly used MER evaluation metrics include Accuracy (ACC) and F1-score. ACC shows the average hit rate across all classes and is susceptible to bias data, and thus it can only reflect the partial effectiveness of an MER classifier. F1-score can remedy the bias issue by computing on the total true positives, false positives and false negatives to reveal the ME classes.

For cross-dataset evaluation, unweighted average recall (UAR) and weighted average recall (WAR) are commonly used~\cite{MEGC2018, zhang2020cross}. WAR refers to the number of correctly classified samples divided by the total number of samples. UAR is defined as mean accuracy of each class divided by the number of classes without consideration of samples per class, which can reduce the bias caused by the class imbalance. 

\subsection{Cross-dataset MER}
Cross-dataset MER is one of the recently emerging while challenging problems in ME analysis. The training sets and testing sets are from different ME datasets, resulting in the inconsistency of feature distributions~\cite{zhang2020multiple}. 

The cross-database recognition in Micro-Expression Grand Challenge (MEGC 2018)~\cite{MEGC2018} used the CASME II and SAMM datasets to serve as the source and target ME database. Zong \textit{et al.}~\cite{zhang2020cross} built a cross-database micro-expression recognition (CDMER) benchmark. There are two types of CDMER tasks. The TYPE-I is conducted between any two subsets of SMIC, i.e., HS, VIS, and NIR. And the TYPE-II uses the selected CASME II and one subset of SMIC, which is proved more difficult than TYPE-I~\cite{zhang2020cross}. The experiments showed the variant of Local Phase Quantization (LPQ), i.e., LPQ-TOP, and LBP with six intersection points (LBP-SIP) performed the best in terms of F1-score and ACC in TYPE-I and TYPE-II. Also, deep features perform rather poorly than most handcrafted features in the CDMER task. The possible reason may be the C3D was pre-trained on Sports1M and UCF101 datasets, whose samples are quite different from ME data. After finetuning the C3D on SMIC (HS), the performance of C3D features can be improved significantly. 
However, with a simple finetuning strategy based on the source database, it seems not enough for learning the database-invariant deep features. 

The heterogeneous problem existing between source and target databases raises the level of difficulty of the CDMER task. For example, SMIC (NIR) is collected by a near-infrared camera, whose image quality is considerably different from images recorded by a high-speed camera (as in CASME II and SMIC (HS)) and visual camera (as in SMIC (VIS)). Li \textit{et al.}~\cite{li2019talsr} proposed the target-adapted least-squares regression (TALSR) to learn a regression coefficient matrix between source samples and the provided labels. The idea is that such coefficient matrix could also reflect the sample-label relation in the target database domain. Zhang \textit{et al.}~\cite{zhang2020multiple} proposed a structure of the super wide regression network (SWiRN) for unsupervised cross-database MER. The \textit{state-of-the-art} domain adaptation methods~\cite{wilson2020survey,zhang2019transfer} can be further exploited to reduce the differences between source and target domains to improve the performance of CDMER methods. 

\subsection{Overall Comparison}

A comparison of MER methodologies with handcrafted features and learning-based features is provided in Table \ref{tab:handcrafted features on CASME II}, Table \ref{tab:handcrafted features on smic} and Table \ref{tab:handcrafted features on samm}, respectively.
Several variants of optical flow and LBP-TOP have also been proposed in recent years. Also, optical flow related features are most frequently used handcrafted features. The experimental results show that the combination of optical flow features and CNN can achieve fine recognition accuracy. Among all methods using handcrafted feature, LEARNet~\cite{verma2019learnet} achieved averagely the best performance, proving the effectiveness of dynamic imaging and the feasibility of using key frames to represent a video sequence.

Although there are many approaches using handcrafed features, we can observe that the trend of MER is changing. In several approaches, low-level handcrafted feature extraction is considered as a pre-stage before high-level feature extraction and fine performances can be obtained. On the other hand, although the end-to-end learning based approaches are reported to have higher performances, they are still restricted by limited labeled data compared to other classification tasks in the literature, encouraging the development of data augmentation or transfer learning works.

While the obtained results from most works are encouraging, there are still some restrictions. Since there is no standard evaluation protocol and class setting, it is hard to reach the conclusions which method performs the best for MER. For example, there are some works only considering three or four emotion labels (i.e., Positive, Negative, Surprise, and Others)~\cite{liong2018biwoof,li2019facial}. The reduction of emotion classes makes the MER task simpler but also introduces the class bias towards negative categories since there is only one positive category (i.e., Happiness). Another problem is that different methods have different dataset splitting or composite strategies~\cite{xia2020revealing,kumar2019classification}. For example, Lai \textit{et al.}~\cite{lai2020real} mixed CASME and CASME II and split the training/testing set as 70/30 percentage. While SMIC, CASME II, and SAMM were combined into a composite dataset in~\cite{kumar2019classification}.

Generally, the problem of real-world automatic micro-expression recognition and spotting still remain challenging, since the existing datasets are collected under the well-controlled environment and the data diversity is very limited. The recognition performances are insufficient and there remain still many topics unexplored.

\section{Challenges and Potential Solutions} \label{sec: Challenges}

\subsection{In-the-wild Scenarios}
Current MER researches have limited scenarios since existing publicly annotated datasets are all collected from lab-controlled environments. In the real-world scenario, more complex and natural emotion including various environmental factors, e.g., illumination interference, 3D head rotation, interaction or interrogation scenarios where more persons involved should be taken into account. 
Lai \textit{et al.}~\cite{lai2020real} took a step forward to avoid excessive computation and established a real-time MER framework with 60 fps running on Intel Xeon 2.10 GHz CPU, 32 GB memory and Ubuntu 14.04 operating system. However, it was experimented on CASME and CASME II, which both are lab-controlled datasets. The MER in-the-wild still remains an open challenge.


\subsection{Uncertainty Modeling}

To figure out why the \textit{state-of-the-art} methods reached a bottleneck in the recognition rate, several works introduced fuzzy set theory~\cite{zadeh1965fuzzy} into MER due to its ability to model the uncertainties. For ambiguous movements of different emotions, Fuzzy Qualitative can model the ambiguity by using the fuzzy membership function to represent each feature dimension with respect to each emotion class. Lim and Goh~\cite{lim2017fuzzy} considered ME as a non-mutual exclusive case. Instead of conducting crisp classification, Fuzzy Qualitative Rank Classifier (FQRC) was proposed to model the ambiguity in MER task by using a multi-label rank classifier. Chen\textit{ et al.}~\cite{chen2016emotion} proposed weighted fuzzy classification for analyzing emotion and achieved promising accuracy. Happy and Routray~\cite{happy2017FHOFO} constructed fuzzy histograms of orientation features based on optical flow.

\subsection{Machine Bias} 
Previous studies have indicated that ingroup advantage in macro-expression recognition exists in various kinds of social groups, e.g., cultural groups, racial groups, and religious groups~\cite{elfenbein2002there,young2018minimal,tuminello2011face,thibault2006effect,young2010mere,huang2014shared,prado2014facial}. 

However, it remains unclear whether the social category of the target influences MER. 
Xie \textit{et al.}~\cite{xie2019ingroup} conducted the intergroup bias experiments among Chinese. The results showed that there is an ingroup disadvantage for the Chinese participants: The recognition accuracy of MEs of outgroup members (White targets) was actually higher than that of ingroup members (Asian targets). And the results also showed that such an intergroup bias is unaffected by the duration of MEs and the ingroup disadvantage remains the same even after the participants had received the training of Micro Expression Training Tool (METT). Regardless of the duration of MEs, there is an ingroup disadvantage for Chinese participants and such a bias still exists even after the training of MER training program. 

One possible solution is to discover the hidden group-invariant features. Investigating the potential factors that may affect the MER help us to develop efficient MER training programs and efficient automatic recognition tools.

\section{Conclusions and Future Research Directions}
\label{sec: Future Recommendation}
The past decade has witnessed the development of many new MER algorithms. This paper provides a comprehensive review of recent advances in MER technology. Future MER works can expand from certain research directions below.

\subsection{Enriching Limited and Biased Datasets}
Current ME databases are usually collected from young subjects, especially undergraduate students. The age range should be extended in the future because MEs decoding is varying among different ages~\cite{folster2014facial}. Meanwhile, the male/female percentage and ethnic groups should also be well considered.
Also, to reduce the labeling process, synthetic data generation algorithms can be further exploited in MER as introduced in Section \ref{sec: Datasets}. 

Due to the subjectivity of human annotators and the ambiguous nature of the expression labels, there might exist annotation inconsistency~\cite{chen2020label,wang2020suppressing}, how to reduce label noises is also needs to consider.

\subsection{Facial Asymmetrical Phenomenon for MER}
\textit{Chirality} is a chemistry term used to describe two objects that appear identical but not symmetrical when folded over onto themselves. Since human faces are naturally asymmetrical~\cite{mandal1995asymmetry}, \textit{facial chirality} demonstrates the asymmetry as the left and the right side of faces differ in emotional communication when people experience multiple emotions at the same time or when there is an attempt to hide an emotion. The general observation is that emotional expressions are more intense on the left side of faces since the right cerebral hemisphere is dominant for the expression~\cite{dopson1984asymmetry}. 
On the other hand, AUs may be coded as symmetrical or asymmetrical. Some existing datasets have left and right AUs annotation that can reveal such phenomenon. Hypothetically, the facial chirality implies that the left side of faces might already include sufficient features that can distinguish one emotion from another. However, there are few MER works extend their researches based on this hypothesis. 

\subsection{Multimodality for MER}
Auxiliary metadata, e.g., words, gestures, voices, can serve as important cues for MER in real world scenarios. Among them, the body gesture is proved to be capable of conveying emotional information.
A \textit{Micro Gesture} dataset~\cite{chen2019analyze} containing 3,692 manually labeled gesture clips was released in 2019. The dataset collects subtle body movements that are elicited when hidden expresions are triggered in unconstrained situations. Their experiments verified the latent relation between one's micro-gestures and the hidden emotional states. Therefore, MEs can be fused with micro-gesture and other physiological signals for a finer level emotion understanding. Moreover, human physiological features, e.g., ECG, EDA, are very informative features for affective revealing, which are less considered in MER works.

\ifCLASSOPTIONcompsoc
  \section*{Acknowledgments}
\else
  \section*{Acknowledgment}
\fi

This work was supported in part by the CTBC Bank under Industry-Academia Cooperation Project and the Ministry of Science and Technology of Taiwan under Grants MOST-108-2218-E-002-055, MOST-109-2223-E-009-002-MY3, MOST-109-2218-E-009-025 and MOST-109-2218-E-002-015.

\ifCLASSOPTIONcaptionsoff
  \newpage
\fi



\bibliographystyle{IEEEtran}
\bibliography{IEEEabrv,reference.bib}

\begin{thebibliography}{100}
\providecommand{\url}[1]{#1}
\csname url@samestyle\endcsname
\providecommand{\newblock}{\relax}
\providecommand{\bibinfo}[2]{#2}
\providecommand{\BIBentrySTDinterwordspacing}{\spaceskip=0pt\relax}
\providecommand{\BIBentryALTinterwordstretchfactor}{4}
\providecommand{\BIBentryALTinterwordspacing}{\spaceskip=\fontdimen2\font plus
\BIBentryALTinterwordstretchfactor\fontdimen3\font minus
  \fontdimen4\font\relax}
\providecommand{\BIBforeignlanguage}[2]{{%
\expandafter\ifx\csname l@#1\endcsname\relax
\typeout{** WARNING: IEEEtran.bst: No hyphenation pattern has been}%
\typeout{** loaded for the language `#1'. Using the pattern for}%
\typeout{** the default language instead.}%
\else
\language=\csname l@#1\endcsname
\fi
#2}}
\providecommand{\BIBdecl}{\relax}
\BIBdecl

\bibitem{ekman2009telling}
P.~Ekman, \emph{Telling lies: Clues to deceit in the marketplace, politics, and
  marriage (revised edition)}.\hskip 1em plus 0.5em minus 0.4em\relax WW Norton
  \& Company, 2009.

\bibitem{MEGC2018}
W.~Merghani, A.~Davison, and M.~Yap, ``Facial micro-expressions grand challenge
  2018: evaluating spatio-temporal features for classification of objective
  classes,'' in \emph{IEEE International Conference on Automatic Face and
  Gesture Recognition}.\hskip 1em plus 0.5em minus 0.4em\relax IEEE, 2018, pp.
  662--666.

\bibitem{MEGC2019}
J.~See, M.~H. Yap, J.~Li, X.~Hong, and S.-J. Wang, ``Megc 2019--the second
  facial micro-expressions grand challenge,'' in \emph{IEEE International
  Conference on Automatic Face and Gesture Recognition}.\hskip 1em plus 0.5em
  minus 0.4em\relax IEEE, 2019, pp. 1--5.

\bibitem{li2020megc2020}
J.~Li, S.~Wang, M.~H. Yap, J.~See, X.~Hong, and X.~Li, ``Megc2020-the third
  facial micro-expression grand challenge,'' in \emph{IEEE International
  Conference on Automatic Face and Gesture Recognition}, 2020, pp. 234--237.

\bibitem{shen2012duration}
X.-b. Shen, Q.~Wu, and X.-l. Fu, ``Effects of the duration of expressions on
  the recognition of microexpressions,'' \emph{Journal of Zhejiang University
  Science B}, vol.~13, no.~3, pp. 221--230, 2012.

\bibitem{yap2019samm}
C.~H. Yap, C.~Kendrick, and M.~H. Yap, ``Samm long videos: A spontaneous facial
  micro-and macro-expressions dataset,'' in \emph{IEEE International Conference
  on Automatic Face and Gesture Recognition}.\hskip 1em plus 0.5em minus
  0.4em\relax IEEE, 2020, pp. 194--199.

\bibitem{yan2013casme}
W.-J. Yan, Q.~Wu, Y.-J. Liu, S.-J. Wang, and X.~Fu, ``Casme database: a dataset
  of spontaneous micro-expressions collected from neutralized faces,'' in
  \emph{IEEE International Conference on Automatic Face and Gesture
  Recognition}.\hskip 1em plus 0.5em minus 0.4em\relax IEEE, 2013, pp. 1--7.

\bibitem{yan2014CASMEII}
W.-J. Yan, X.~Li, S.-J. Wang, G.~Zhao, Y.-J. Liu, Y.-H. Chen, and X.~Fu,
  ``Casme ii: An improved spontaneous micro-expression database and the
  baseline evaluation,'' \emph{PloS one}, vol.~9, no.~1, p. e86041, 2014.

\bibitem{shen2016electrophysiological}
X.~Shen, Q.~Wu, K.~Zhao, and X.~Fu, ``Electrophysiological evidence reveals
  differences between the recognition of microexpressions and
  macroexpressions,'' \emph{Frontiers in psychology}, vol.~7, p. 1346, 2016.

\bibitem{ekman1997face}
R.~Ekman, \emph{What the face reveals: Basic and applied studies of spontaneous
  expression using the Facial Action Coding System (FACS)}.\hskip 1em plus
  0.5em minus 0.4em\relax Oxford University Press, USA, 1997.

\bibitem{rosenberg2020face}
E.~L. Rosenberg and P.~Ekman, \emph{What the face reveals: Basic and applied
  studies of spontaneous expression using the facial action coding system
  (FACS)}.\hskip 1em plus 0.5em minus 0.4em\relax Oxford University Press,
  2020.

\bibitem{wang2015micro}
S.-J. Wang, W.-J. Yan, X.~Li, G.~Zhao, C.-G. Zhou, X.~Fu, M.~Yang, and J.~Tao,
  ``Micro-expression recognition using color spaces,'' \emph{IEEE Transactions
  on Image Processing}, vol.~24, no.~12, pp. 6034--6047, 2015.

\bibitem{pantic2009machine}
M.~Pantic, ``Machine analysis of facial behaviour: Naturalistic and dynamic
  behaviour,'' \emph{Philosophical Transactions of the Royal Society B:
  Biological Sciences}, vol. 364, no. 1535, pp. 3505--3513, 2009.

\bibitem{EMFACS7}
W.~V. Friesen and P.~Ekman, \emph{EMFACS-7: Emotional Facial Action Coding
  System, Version 7}.\hskip 1em plus 0.5em minus 0.4em\relax Unpublished
  manual, 1984.

\bibitem{ekman2002facial}
P.~Ekman, W.~V. Friesen, and J.~C. Hager, ``Facial action coding system: The
  manual on cd rom,'' \emph{A Human Face, Salt Lake City}, pp. 77--254, 2002.

\bibitem{FACSAID}
R.~Cairns, G.~Elder~Jr, and E.~Costello, ``Cambridge studies in social and
  emotional development,'' \emph{Developmental science. Cambridge University
  Press}, 1996.

\bibitem{duran2017coherence}
J.~I. Dur{\'a}n, R.~Reisenzein, and J.-M. Fern{\'a}ndez-Dols, ``Coherence
  between emotions and facial expressions,'' \emph{The science of facial
  expression}, pp. 107--129, 2017.

\bibitem{yan2014micro}
W.-J. Yan, S.-J. Wang, Y.-J. Liu, Q.~Wu, and X.~Fu, ``For micro-expression
  recognition: Database and suggestions,'' \emph{Neurocomputing}, vol. 136, pp.
  82--87, 2014.

\bibitem{li2017towards}
X.~Li, X.~Hong, A.~Moilanen, X.~Huang, T.~Pfister, G.~Zhao, and
  M.~Pietik{\"a}inen, ``Towards reading hidden emotions: A comparative study of
  spontaneous micro-expression spotting and recognition methods,'' \emph{IEEE
  Transactions on Affective Computing}, vol.~9, no.~4, pp. 563--577, 2017.

\bibitem{goh2018micro}
K.~M. Goh, C.~H. Ng, L.~L. Lim, and U.~U. Sheikh, ``Micro-expression
  recognition: an updated review of current trends, challenges and solutions,''
  \emph{The Visual Computer}, vol.~36, no.~3, pp. 445--468, 2018.

\bibitem{oh2018survey}
Y.-H. Oh, J.~See, A.~C. Le~Ngo, R.~C.-W. Phan, and V.~M. Baskaran, ``A survey
  of automatic facial micro-expression analysis: databases, methods, and
  challenges,'' \emph{Frontiers in Psychology}, vol.~9, p. 1128, 2018.

\bibitem{merghani2018review}
W.~Merghani, A.~K. Davison, and M.~H. Yap, ``A review on facial
  micro-expressions analysis: datasets, features and metrics,'' \emph{arXiv
  preprint arXiv:1805.02397}, 2018.

\bibitem{takalkar2018survey}
M.~Takalkar, M.~Xu, Q.~Wu, and Z.~Chaczko, ``A survey: facial micro-expression
  recognition,'' \emph{Multimedia Tools and Applications}, vol.~77, no.~15, pp.
  19\,301--19\,325, 2018.

\bibitem{zhou2020survey}
L.~Zhou, X.~Shao, and Q.~Mao, ``A survey of micro-expression recognition,''
  \emph{Image and Vision Computing}, p. 104043, 2020.

\bibitem{ekman1992_emotionclass}
P.~Ekman, ``Facial expressions of emotion: an old controversy and new
  findings,'' \emph{Philosophical Transactions of the Royal Society of London.
  Series B: Biological Sciences}, vol. 335, no. 1273, pp. 63--69, 1992.

\bibitem{SMIC}
X.~Li, T.~Pfister, X.~Huang, G.~Zhao, and M.~Pietik{\"a}inen, ``A spontaneous
  micro-expression database: Inducement, collection and baseline,'' in
  \emph{IEEE International Conference on Automatic Face and Gesture
  Recognition}.\hskip 1em plus 0.5em minus 0.4em\relax IEEE, 2013, pp. 1--6.

\bibitem{davison2018objective}
A.~K. Davison, W.~Merghani, and M.~H. Yap, ``Objective classes for micro-facial
  expression recognition,'' \emph{Journal of Imaging}, vol.~4, no.~10, p. 119,
  2018.

\bibitem{merghani2019implication}
W.~Merghani, A.~K. Davison, and M.~H. Yap, ``The implication of spatial
  temporal changes on facial micro-expression analysis,'' \emph{Multimedia
  Tools and Applications}, vol.~78, no.~15, pp. 21\,613--21\,628, 2019.

\bibitem{coan2007-EmotionElicitation}
J.~A. Coan and J.~J. Allen, \emph{Handbook of emotion elicitation and
  assessment}.\hskip 1em plus 0.5em minus 0.4em\relax Oxford university press,
  2007.

\bibitem{lang1980behavioral_SAM}
P.~Lang, ``Behavioral treatment and bio-behavioral assessment: Computer
  applications,'' \emph{Technology in Mental Health Care Delivery Systems}, pp.
  119--137, 1980.

\bibitem{ekman2009lie}
P.~Ekman, ``Lie catching and microexpressions,'' \emph{The Philosophy of
  Deception}, vol.~1, no.~2, p.~5, 2009.

\bibitem{xie2020assisted}
H.-X. Xie, L.~Lo, H.-H. Shuai, and W.-H. Cheng, ``Au-assisted graph attention
  convolutional network for micro-expression recognition,'' in
  \emph{Proceedings of the 28th ACM International Conference on
  Multimedia}.\hskip 1em plus 0.5em minus 0.4em\relax ACM, 2020, pp.
  2871--2880.

\bibitem{qu2017cas}
F.~Qu, S.-J. Wang, W.-J. Yan, H.~Li, S.~Wu, and X.~Fu, ``Cas (me) $^2$: A
  database for spontaneous macro-expression and micro-expression spotting and
  recognition,'' \emph{IEEE Transactions on Affective Computing}, vol.~9,
  no.~4, pp. 424--436, 2017.

\bibitem{frank2009see}
M.~Frank, M.~Herbasz, K.~Sinuk, A.~Keller, and C.~Nolan, ``I see how you feel:
  Training laypeople and professionals to recognize fleeting emotions,'' in
  \emph{The Annual Meeting of the International Communication
  Association.}\hskip 1em plus 0.5em minus 0.4em\relax ICA, 2009, pp. 1--35.

\bibitem{freeman2000learning}
W.~T. Freeman, E.~C. Pasztor, and O.~T. Carmichael, ``Learning low-level
  vision,'' \emph{International Journal of Computer Vision}, vol.~40, no.~1,
  pp. 25--47, 2000.

\bibitem{butler2012naturalistic}
D.~J. Butler, J.~Wulff, G.~B. Stanley, and M.~J. Black, ``A naturalistic open
  source movie for optical flow evaluation,'' in \emph{European Conference on
  Computer Vision}.\hskip 1em plus 0.5em minus 0.4em\relax Springer, 2012, pp.
  611--625.

\bibitem{dosovitskiy2015flownet}
A.~Dosovitskiy, P.~Fischer, E.~Ilg, P.~Hausser, C.~Hazirbas, V.~Golkov, P.~Van
  Der~Smagt, D.~Cremers, and T.~Brox, ``Flownet: Learning optical flow with
  convolutional networks,'' in \emph{IEEE International Conference on Computer
  Vision}.\hskip 1em plus 0.5em minus 0.4em\relax IEEE, 2015, pp. 2758--2766.

\bibitem{wang2019learning}
Q.~Wang, J.~Gao, W.~Lin, and Y.~Yuan, ``Learning from synthetic data for crowd
  counting in the wild,'' in \emph{IEEE International Conference on Computer
  Vision and Pattern Recognition}.\hskip 1em plus 0.5em minus 0.4em\relax IEEE,
  2019, pp. 8198--8207.

\bibitem{richter2016playing}
S.~R. Richter, V.~Vineet, S.~Roth, and V.~Koltun, ``Playing for data: Ground
  truth from computer games,'' in \emph{European Conference on Computer
  Vision}.\hskip 1em plus 0.5em minus 0.4em\relax Springer, 2016, pp. 102--118.

\bibitem{papon2015semantic}
J.~Papon and M.~Schoeler, ``Semantic pose using deep networks trained on
  synthetic rgb-d,'' in \emph{IEEE International Conference on Computer
  Vision}.\hskip 1em plus 0.5em minus 0.4em\relax IEEE, 2015, pp. 774--782.

\bibitem{9229137}
G.~J. {Qi} and J.~{Luo}, ``Small data challenges in big data era: A survey of
  recent progress on unsupervised and semi-supervised methods,'' \emph{IEEE
  Transactions on Pattern Analysis and Machine Intelligence}, pp. 1--1, 2020.

\bibitem{queiroz2010generating}
R.~Queiroz, M.~Cohen, J.~L. Moreira, A.~Braun, J.~C.~J. J{\'u}nior, and S.~R.
  Musse, ``Generating facial ground truth with synthetic faces,'' in \emph{IEEE
  Conference on Graphics, Patterns and Images}.\hskip 1em plus 0.5em minus
  0.4em\relax IEEE, 2010, pp. 25--31.

\bibitem{abbasnejad2017using}
I.~Abbasnejad, S.~Sridharan, D.~Nguyen, S.~Denman, C.~Fookes, and S.~Lucey,
  ``Using synthetic data to improve facial expression analysis with 3d
  convolutional networks,'' in \emph{IEEE International Conference on Computer
  Vision Workshops}.\hskip 1em plus 0.5em minus 0.4em\relax IEEE, 2017, pp.
  1609--1618.

\bibitem{zhang2018joint}
F.~Zhang, T.~Zhang, Q.~Mao, and C.~Xu, ``Joint pose and expression modeling for
  facial expression recognition,'' in \emph{IEEE International Conference on
  Computer Vision and Pattern Recognition}.\hskip 1em plus 0.5em minus
  0.4em\relax IEEE, 2018, pp. 3359--3368.

\bibitem{cai2019identity}
J.~Cai, Z.~Meng, A.~S. Khan, Z.~Li, J.~O'Reilly, and Y.~Tong, ``Identity-free
  facial expression recognition using conditional generative adversarial
  network,'' \emph{arXiv preprint arXiv:1903.08051}, 2019.

\bibitem{yu2019lcbp}
M.~Yu, Z.~Guo, Y.~Yu, Y.~Wang, and S.~Cen, ``Spatiotemporal feature descriptor
  for micro-expression recognition using local cube binary pattern,''
  \emph{IEEE Access}, vol.~7, pp. 159\,214--159\,225, 2019.

\bibitem{hu2019gender}
C.~Hu, J.~Chen, X.~Zuo, H.~Zou, X.~Deng, and Y.~Shu, ``Gender-specific
  multi-task micro-expression recognition using pyramid cgbp-top feature,''
  \emph{Computer Modeling in Engineering and Sciences}, vol. 118, no.~3, pp.
  547--559, 2019.

\bibitem{carcagni2015facial}
P.~Carcagn{\`\i}, M.~Del~Coco, M.~Leo, and C.~Distante, ``Facial expression
  recognition and histograms of oriented gradients: a comprehensive study,''
  \emph{SpringerPlus}, vol.~4, no.~1, p. 645, 2015.

\bibitem{niu2018LTOGP}
M.~Niu, Y.~Li, J.~Tao, and S.-J. Wang, ``Micro-expression recognition based on
  local two-order gradient pattern,'' in \emph{IEEE Asian Conference on
  Affective Computing and Intelligent Interaction}.\hskip 1em plus 0.5em minus
  0.4em\relax IEEE, 2018, pp. 1--6.

\bibitem{liong2018biwoof}
S.-T. Liong, J.~See, K.~Wong, and R.~C.-W. Phan, ``Less is more:
  Micro-expression recognition from video using apex frame,'' \emph{Signal
  Processing: Image Communication}, vol.~62, pp. 82--92, 2018.

\bibitem{sun2020dynamic}
B.~Sun, S.~Cao, D.~Li, J.~He, and L.~Yu, ``Dynamic micro-expression recognition
  using knowledge distillation,'' \emph{IEEE Transactions on Affective
  Computing}, 2020.

\bibitem{huang2017discriminative}
X.~Huang, S.-J. Wang, X.~Liu, G.~Zhao, X.~Feng, and M.~Pietik{\"a}inen,
  ``Discriminative spatiotemporal local binary pattern with revisited integral
  projection for spontaneous facial micro-expression recognition,'' \emph{IEEE
  Transactions on Affective Computing}, vol.~10, no.~1, pp. 32--47, 2017.

\bibitem{viola2001rapid}
P.~Viola and M.~Jones, ``Rapid object detection using a boosted cascade of
  simple features,'' in \emph{IEEE International Conference on Computer Vision
  and Pattern Recognition}, vol.~1.\hskip 1em plus 0.5em minus 0.4em\relax
  IEEE, 2001, pp. I--I.

\bibitem{dalal2005histograms}
N.~Dalal and B.~Triggs, ``Histograms of oriented gradients for human
  detection,'' in \emph{IEEE International Conference on Computer Vision and
  Pattern Recognition}.\hskip 1em plus 0.5em minus 0.4em\relax IEEE, 2005, pp.
  886--893.

\bibitem{sanchez2017robust}
J.~Sanchez-Riera, K.~Srinivasan, K.-L. Hua, W.-H. Cheng, M.~A. Hossain, and
  M.~F. Alhamid, ``Robust rgb-d hand tracking using deep learning priors,''
  \emph{IEEE Transactions on Circuits and Systems for Video Technology},
  vol.~28, no.~9, pp. 2289--2301, 2017.

\bibitem{king2015max}
D.~E. King, ``Max-margin object detection,'' \emph{arXiv preprint
  arXiv:1502.00046}, 2015.

\bibitem{masi2018deep}
I.~Masi, Y.~Wu, T.~Hassner, and P.~Natarajan, ``Deep face recognition: A
  survey,'' in \emph{2018 31st SIBGRAPI conference on graphics, patterns and
  images (SIBGRAPI)}.\hskip 1em plus 0.5em minus 0.4em\relax IEEE, 2018, pp.
  471--478.

\bibitem{zhang2020spatio}
L.-w. Zhang, J.~Li, S.~Wang, X.~Duan, W.~Yan, H.~Xie, and S.~Huang,
  ``Spatio-temporal fusion for macro-and micro-expression spotting in long
  video sequences,'' in \emph{IEEE International Conference on Automatic Face
  and Gesture Recognition}.\hskip 1em plus 0.5em minus 0.4em\relax IEEE, 2020,
  pp. 245--252.

\bibitem{verburg2019micro}
M.~Verburg and V.~Menkovski, ``Micro-expression detection in long videos using
  optical flow and recurrent neural networks,'' in \emph{IEEE International
  Conference on Automatic Face and Gesture Recognition}.\hskip 1em plus 0.5em
  minus 0.4em\relax IEEE, 2019, pp. 1--6.

\bibitem{li2019spotting}
J.~Li, C.~Soladie, R.~Seguier, S.-J. Wang, and M.~H. Yap, ``Spotting
  micro-expressions on long videos sequences,'' in \emph{IEEE International
  Conference on Automatic Face and Gesture Recognition}.\hskip 1em plus 0.5em
  minus 0.4em\relax IEEE, 2019, pp. 1--5.

\bibitem{tran2019dense}
T.-K. Tran, Q.-N. Vo, X.~Hong, and G.~Zhao, ``Dense prediction for
  micro-expression spotting based on deep sequence model,'' \emph{Electronic
  Imaging}, vol. 2019, no.~8, pp. 401--1, 2019.

\bibitem{beh2019micro}
K.~X. Beh and K.~M. Goh, ``Micro-expression spotting using facial landmarks,''
  in \emph{IEEE International Colloquium on Signal Processing and Its
  Applications}.\hskip 1em plus 0.5em minus 0.4em\relax IEEE, 2019, pp.
  192--197.

\bibitem{nag2019facial}
S.~Nag, A.~K. Bhunia, A.~Konwer, and P.~P. Roy, ``Facial micro-expression
  spotting and recognition using time contrasted feature with visual memory,''
  in \emph{IEEE International Conference on Acoustics, Speech and Signal
  Processing}.\hskip 1em plus 0.5em minus 0.4em\relax IEEE, 2019, pp.
  2022--2026.

\bibitem{han2018cfd}
Y.~Han, B.~Li, Y.-K. Lai, and Y.-J. Liu, ``Cfd: a collaborative feature
  difference method for spontaneous micro-expression spotting,'' in \emph{IEEE
  International Conference on Image Processing}.\hskip 1em plus 0.5em minus
  0.4em\relax IEEE, 2018, pp. 1942--1946.

\bibitem{duque2018micro}
C.~A. Duque, O.~Alata, R.~Emonet, A.-C. Legrand, and H.~Konik,
  ``Micro-expression spotting using the riesz pyramid,'' in \emph{IEEE Winter
  Conference on Applications of Computer Vision}.\hskip 1em plus 0.5em minus
  0.4em\relax IEEE, 2018, pp. 66--74.

\bibitem{milborrow2008locating}
S.~Milborrow and F.~Nicolls, ``Locating facial features with an extended active
  shape model,'' in \emph{European conference on computer vision}.\hskip 1em
  plus 0.5em minus 0.4em\relax Springer, 2008, pp. 504--513.

\bibitem{asthana2013robust}
A.~Asthana, S.~Zafeiriou, S.~Cheng, and M.~Pantic, ``Robust discriminative
  response map fitting with constrained local models,'' in \emph{IEEE
  Conference on Computer Vision and Pattern Recognition}.\hskip 1em plus 0.5em
  minus 0.4em\relax IEEE, 2013, pp. 3444--3451.

\bibitem{wu2017facial}
Y.~Wu, T.~Hassner, K.~Kim, G.~Medioni, and P.~Natarajan, ``Facial landmark
  detection with tweaked convolutional neural networks,'' \emph{IEEE
  Transactions on Pattern Analysis and Machine Intelligence}, vol.~40, no.~12,
  pp. 3067--3074, 2017.

\bibitem{birchfield1997derivation}
S.~Birchfield, ``Derivation of kanade-lucas-tomasi tracking equation,''
  \emph{unpublished notes}, 1997.

\bibitem{le2019seeing}
A.~C. Le~Ngo and R.~C.-W. Phan, ``Seeing the invisible: Survey of video motion
  magnification and small motion analysis,'' \emph{ACM Computing Surveys
  (CSUR)}, vol.~52, no.~6, pp. 1--20, 2019.

\bibitem{le2018micro}
A.~C. Le~Ngo, A.~Johnston, R.~C.-W. Phan, and J.~See, ``Micro-expression motion
  magnification: Global lagrangian vs. local eulerian approaches,'' in
  \emph{IEEE International Conference on Automatic Face and Gesture
  Recognition}.\hskip 1em plus 0.5em minus 0.4em\relax IEEE, 2018, pp.
  650--656.

\bibitem{mm_lei2020}
L.~Lei, J.~Li, T.~Chen, and S.~Li, ``A novel graph-tcn with a graph structured
  representation for micro-expression recognition,'' in \emph{Proceedings of
  the 28th ACM International Conference on Multimedia}.\hskip 1em plus 0.5em
  minus 0.4em\relax ACM, 2020, pp. 2237--2245.

\bibitem{zeng2008survey}
Z.~Zeng, M.~Pantic, G.~I. Roisman, and T.~S. Huang, ``A survey of affect
  recognition methods: Audio, visual, and spontaneous expressions,'' \emph{IEEE
  Transactions on Pattern Analysis and Machine Antelligence}, vol.~31, no.~1,
  pp. 39--58, 2008.

\bibitem{lin2012human}
Y.-C. Lin, M.-C. Hu, W.-H. Cheng, Y.-H. Hsieh, and H.-M. Chen, ``Human action
  recognition and retrieval using sole depth information,'' in
  \emph{Proceedings of the 20th ACM international conference on Multimedia},
  2012, pp. 1053--1056.

\bibitem{hu2014real}
M.-C. Hu, C.-W. Chen, W.-H. Cheng, C.-H. Chang, J.-H. Lai, and J.-L. Wu,
  ``Real-time human movement retrieval and assessment with kinect sensor,''
  \emph{IEEE transactions on cybernetics}, vol.~45, no.~4, pp. 742--753, 2014.

\bibitem{shen2015gestalt}
I.-C. Shen and W.-H. Cheng, ``Gestalt rule feature points,'' \emph{IEEE
  Transactions on Multimedia}, vol.~17, no.~4, pp. 526--537, 2015.

\bibitem{rivera2012local}
A.~R. Rivera, J.~R. Castillo, and O.~O. Chae, ``Local directional number
  pattern for face analysis: Face and expression recognition,'' \emph{IEEE
  transactions on image processing}, vol.~22, no.~5, pp. 1740--1752, 2012.

\bibitem{ojala1994lbp}
T.~Ojala, M.~Pietikainen, and D.~Harwood, ``Performance evaluation of texture
  measures with classification based on kullback discrimination of
  distributions,'' in \emph{IEEE International Conference on Computer Vision
  and Pattern Recognition}.\hskip 1em plus 0.5em minus 0.4em\relax IEEE, 1994,
  pp. 582--585.

\bibitem{ojala2002multiresolution}
T.~Ojala, M.~Pietikainen, and T.~Maenpaa, ``Multiresolution gray-scale and
  rotation invariant texture classification with local binary patterns,''
  \emph{IEEE Transactions on Pattern Analysis and Machine Intelligence},
  vol.~24, no.~7, pp. 971--987, 2002.

\bibitem{zhao2007dynamic}
G.~Zhao and M.~Pietikainen, ``Dynamic texture recognition using local binary
  patterns with an application to facial expressions,'' \emph{IEEE Transactions
  on Pattern Analysis and Machine Intelligence}, vol.~29, no.~6, pp. 915--928,
  2007.

\bibitem{zong2018domain}
Y.~Zong, W.~Zheng, X.~Huang, J.~Shi, Z.~Cui, and G.~Zhao, ``Domain regeneration
  for cross-database micro-expression recognition,'' \emph{IEEE Transactions on
  Image Processing}, vol.~27, no.~5, pp. 2484--2498, 2018.

\bibitem{zhang2020new}
Y.~Zhang, H.~Jiang, X.~Li, B.~Lu, K.~M. Rabie, and A.~U. Rehman, ``A new
  framework combining local-region division and feature selection for
  micro-expressions recognition,'' \emph{IEEE Access}, vol.~8, pp.
  94\,499--94\,509, 2020.

\bibitem{mm_xia2020macro}
B.~Xia, W.~Wang, S.~Wang, and E.~Chen, ``Learning from macro-expression: a
  micro-expression recognition framework,'' in \emph{Proceedings of the 28th
  ACM International Conference on Multimedia}.\hskip 1em plus 0.5em minus
  0.4em\relax ACM, 2020, pp. 2936--2944.

\bibitem{zhi2019combining}
R.~Zhi, H.~Xu, M.~Wan, and T.~Li, ``Combining 3d convolutional neural networks
  with transfer learning by supervised pre-training for facial micro-expression
  recognition,'' \emph{IEICE Transactions on Information and Systems}, vol.
  102, no.~5, pp. 1054--1064, 2019.

\bibitem{zhou2019cross}
L.~Zhou, Q.~Mao, and L.~Xue, ``Cross-database micro-expression recognition: a
  style aggregated and attention transfer approach,'' in \emph{IEEE
  International Conference on Multimedia and Expo Workshops}.\hskip 1em plus
  0.5em minus 0.4em\relax IEEE, 2019, pp. 102--107.

\bibitem{jia2018macro}
X.~Jia, X.~Ben, H.~Yuan, K.~Kpalma, and W.~Meng, ``Macro-to-micro
  transformation model for micro-expression recognition,'' \emph{Journal of
  Computational Science}, vol.~25, pp. 289--297, 2018.

\bibitem{wang2020micro}
C.~Wang, M.~Peng, T.~Bi, and T.~Chen, ``Micro-attention for micro-expression
  recognition,'' \emph{Neurocomputing}, vol. 410, pp. 354--362, 2020.

\bibitem{peng2018macro}
M.~Peng, Z.~Wu, Z.~Zhang, and T.~Chen, ``From macro to micro expression
  recognition: Deep learning on small datasets using transfer learning,'' in
  \emph{IEEE International Conference on Automatic Face and Gesture
  Recognition}.\hskip 1em plus 0.5em minus 0.4em\relax IEEE, 2018, pp.
  657--661.

\bibitem{ben2018learning}
X.~Ben, X.~Jia, R.~Yan, X.~Zhang, and W.~Meng, ``Learning effective binary
  descriptors for micro-expression recognition transferred by
  macro-information,'' \emph{Pattern Recognition Letters}, vol. 107, pp.
  50--58, 2018.

\bibitem{mao2019classroom}
L.~Mao, N.~Wang, L.~Wang, and Y.~Chen, ``Classroom micro-expression recognition
  algorithms based on multi-feature fusion,'' \emph{IEEE Access}, vol.~7, pp.
  64\,978--64\,983, 2019.

\bibitem{guo2019elbptop}
C.~Guo, J.~Liang, G.~Zhan, Z.~Liu, M.~Pietik{\"a}inen, and L.~Liu, ``Extended
  local binary patterns for efficient and robust spontaneous facial
  micro-expression recognition,'' \emph{IEEE Access}, vol.~7, pp.
  174\,517--174\,530, 2019.

\bibitem{hu2018multi}
C.~Hu, D.~Jiang, H.~Zou, X.~Zuo, and Y.~Shu, ``Multi-task micro-expression
  recognition combining deep and handcrafted features,'' in \emph{IEEE
  International Conference on Pattern Recognition)}.\hskip 1em plus 0.5em minus
  0.4em\relax IEEE, 2018, pp. 946--951.

\bibitem{arango2020mean}
C.~Arango~Duque, O.~Alata, R.~Emonet, H.~Konik, and A.-C. Legrand, ``Mean
  oriented riesz features for micro expression classification,'' \emph{arXiv},
  pp. arXiv--2005, 2020.

\bibitem{wu2016unfolding}
B.~Wu, T.~Mei, W.-H. Cheng, and Y.~Zhang, ``Unfolding temporal dynamics:
  Predicting social media popularity using multi-scale temporal
  decomposition,'' 2016.

\bibitem{wu2016time}
B.~Wu, W.-H. Cheng, Y.~Zhang, and T.~Mei, ``Time matters: Multi-scale
  temporalization of social media popularity,'' in \emph{Proceedings of the
  24th ACM international conference on Multimedia}, 2016, pp. 1336--1344.

\bibitem{horn1981determining}
B.~K. Horn and B.~G. Schunck, ``Determining optical flow,'' in \emph{Techniques
  and Applications of Image Understanding}, vol. 281.\hskip 1em plus 0.5em
  minus 0.4em\relax International Society for Optics and Photonics, 1981, pp.
  319--331.

\bibitem{xia2020revealing}
Z.~Xia, W.~Peng, H.-Q. Khor, X.~Feng, and G.~Zhao, ``Revealing the invisible
  with model and data shrinking for composite-database micro-expression
  recognition,'' \emph{IEEE Transactions on Image Processing}, vol.~29, pp.
  8590--8605, 2020.

\bibitem{liong2019shallow}
S.-T. Liong, Y.~Gan, J.~See, H.-Q. Khor, and Y.-C. Huang, ``Shallow triple
  stream three-dimensional cnn (ststnet) for micro-expression recognition,'' in
  \emph{IEEE International Conference on Automatic Face and Gesture
  Recognition}.\hskip 1em plus 0.5em minus 0.4em\relax IEEE, 2019, pp. 1--5.

\bibitem{gan2019offapexnet}
Y.~Gan, S.-T. Liong, W.-C. Yau, Y.-C. Huang, and L.-K. Tan, ``Off-apexnet on
  micro-expression recognition system,'' \emph{Signal Processing: Image
  Communication}, vol.~74, pp. 129--139, 2019.

\bibitem{liu2019neural}
Y.~Liu, H.~Du, L.~Zheng, and T.~Gedeon, ``A neural micro-expression
  recognizer,'' in \emph{IEEE International Conference on Automatic Face and
  Gesture Recognition}.\hskip 1em plus 0.5em minus 0.4em\relax IEEE, 2019, pp.
  1--4.

\bibitem{zhao2019convolutional}
Y.~Zhao and J.~Xu, ``A convolutional neural network for compound
  micro-expression recognition,'' \emph{Sensors}, vol.~19, no.~24, p. 5553,
  2019.

\bibitem{gan2018bi}
Y.~Gan and S.-T. Liong, ``Bi-directional vectors from apex in cnn for
  micro-expression recognition,'' in \emph{IEEE International Conference on
  Image, Vision and Computing}.\hskip 1em plus 0.5em minus 0.4em\relax IEEE,
  2018, pp. 168--172.

\bibitem{li2019facial}
Q.~Li, S.~Zhan, L.~Xu, and C.~Wu, ``Facial micro-expression recognition based
  on the fusion of deep learning and enhanced optical flow,'' \emph{Multimedia
  Tools and Applications}, vol.~78, no.~20, pp. 29\,307--29\,322, 2019.

\bibitem{li2018fusing}
Q.~Li, J.~Yu, T.~Kurihara, and S.~Zhan, ``Micro-expression analysis by fusing
  deep convolutional neural network and optical flow,'' in \emph{IEEE
  International Conference on Control, Decision and Information
  Technologies)}.\hskip 1em plus 0.5em minus 0.4em\relax IEEE, 2018, pp.
  265--270.

\bibitem{happy2017FHOFO}
S.~Happy and A.~Routray, ``Fuzzy histogram of optical flow orientations for
  micro-expression recognition,'' \emph{IEEE Transactions on Affective
  Computing}, vol.~10, no.~3, pp. 394--406, 2017.

\bibitem{li2018ltp}
J.~Li, C.~Soladie, and R.~Seguier, ``Ltp-ml: Micro-expression detection by
  recognition of local temporal pattern of facial movements,'' in \emph{IEEE
  International Conference on Automatic Face and Gesture Recognition}.\hskip
  1em plus 0.5em minus 0.4em\relax IEEE, 2018, pp. 634--641.

\bibitem{pawar2019micro}
S.~S. Pawar, M.~Moh, and T.-S. Moh, ``Micro-expression recognition using motion
  magnification and spatiotemporal texture map,'' in \emph{International
  Conference on Ubiquitous Information Management and Communication}.\hskip 1em
  plus 0.5em minus 0.4em\relax Springer, 2019, pp. 351--369.

\bibitem{sanchez2016comparative}
J.~Sanchez-Riera, K.-L. Hua, Y.-S. Hsiao, T.~Lim, S.~C. Hidayati, and W.-H.
  Cheng, ``A comparative study of data fusion for rgb-d based visual
  recognition,'' \emph{Pattern Recognition Letters}, vol.~73, pp. 1--6, 2016.

\bibitem{lin2018gabor}
C.~Lin, F.~Long, J.~Huang, and J.~Li, ``Micro-expression recognition based on
  spatiotemporal gabor filters,'' in \emph{IEEE International Conference on
  Information Science and Technology}.\hskip 1em plus 0.5em minus 0.4em\relax
  IEEE, 2018, pp. 487--491.

\bibitem{zong2018hierarchical}
Y.~Zong, X.~Huang, W.~Zheng, Z.~Cui, and G.~Zhao, ``Learning from hierarchical
  spatiotemporal descriptors for micro-expression recognition,'' \emph{IEEE
  Transactions on Multimedia}, vol.~20, no.~11, pp. 3160--3172, 2018.

\bibitem{wang2019weighted}
L.~Wang, H.~Xiao, S.~Luo, J.~Zhang, and X.~Liu, ``A weighted feature extraction
  method based on temporal accumulation of optical flow for micro-expression
  recognition,'' \emph{Signal Processing: Image Communication}, vol.~78, pp.
  246--253, 2019.

\bibitem{zhao2019improved}
Y.~Zhao and J.~Xu, ``An improved micro-expression recognition method based on
  necessary morphological patches,'' \emph{Symmetry}, vol.~11, no.~4, p. 497,
  2019.

\bibitem{zhong2020facial}
W.~Zhong, X.~Yu, L.~Shi, and Z.~Xie, ``Facial micro-expression recognition
  based on local region of the key frame,'' in \emph{International Symposium on
  Multispectral Image Processing and Pattern Recognition}, vol. 11430.\hskip
  1em plus 0.5em minus 0.4em\relax International Society for Optics and
  Photonics, 2020, p. 114301L.

\bibitem{gan2019off}
Y.~Gan, S.-T. Liong, W.-C. Yau, Y.-C. Huang, and L.-K. Tan, ``Off-apexnet on
  micro-expression recognition system,'' \emph{Signal Processing: Image
  Communication}, vol.~74, pp. 129--139, 2019.

\bibitem{liong2018less}
S.-T. Liong, J.~See, K.~Wong, and R.~C.-W. Phan, ``Less is more:
  Micro-expression recognition from video using apex frame,'' \emph{Signal
  Processing: Image Communication}, vol.~62, pp. 82--92, 2018.

\bibitem{deng2009imagenet}
J.~Deng, W.~Dong, R.~Socher, L.-J. Li, K.~Li, and L.~Fei-Fei, ``Imagenet: A
  large-scale hierarchical image database,'' in \emph{IEEE conference on
  Computer Vision and Pattern Recognition}.\hskip 1em plus 0.5em minus
  0.4em\relax IEEE, 2009, pp. 248--255.

\bibitem{ren2015faster}
S.~Ren, K.~He, R.~Girshick, and J.~Sun, ``Faster r-cnn: Towards real-time
  object detection with region proposal networks,'' in \emph{Advances in neural
  information processing systems}, 2015, pp. 91--99.

\bibitem{wang2017background}
H.-C. Wang, Y.-C. Lai, W.-H. Cheng, C.-Y. Cheng, and K.-L. Hua, ``Background
  extraction based on joint gaussian conditional random fields,'' \emph{IEEE
  Transactions on Circuits and Systems for Video Technology}, vol.~28, no.~11,
  pp. 3127--3140, 2017.

\bibitem{hua2015computer}
K.-L. Hua, C.-H. Hsu, S.~C. Hidayati, W.-H. Cheng, and Y.-J. Chen,
  ``Computer-aided classification of lung nodules on computed tomography images
  via deep learning technique,'' \emph{OncoTargets and therapy}, vol.~8, 2015.

\bibitem{abousaleh2016novel}
F.~S. Abousaleh, T.~Lim, W.-H. Cheng, N.-H. Yu, M.~A. Hossain, and M.~F.
  Alhamid, ``A novel comparative deep learning framework for facial age
  estimation,'' \emph{EURASIP Journal on Image and Video Processing}, vol.
  2016, no.~1, p.~47, 2016.

\bibitem{hsieh2019fashionon}
C.-W. Hsieh, C.-Y. Chen, C.-L. Chou, H.-H. Shuai, J.~Liu, and W.-H. Cheng,
  ``Fashionon: Semantic-guided image-based virtual try-on with detailed human
  and clothing information,'' in \emph{Proceedings of the 27th ACM
  International Conference on Multimedia}.\hskip 1em plus 0.5em minus
  0.4em\relax ACM, 2019, pp. 275--283.

\bibitem{susanto2020emotion}
I.~Y. Susanto, T.-Y. Pan, C.-W. Chen, M.-C. Hu, and W.-H. Cheng, ``Emotion
  recognition from galvanic skin response signal based on deep hybrid neural
  networks,'' in \emph{International Conference on Multimedia Retrieval}.\hskip
  1em plus 0.5em minus 0.4em\relax ACM, 2020, pp. 341--345.

\bibitem{cheng2018ai+}
W.-H. Cheng, J.~Liu, M.~S. Kankanhalli, A.~El~Saddik, and B.~Huet, ``Ai+
  multimedia make better life?'' in \emph{Proceedings of the 26th ACM
  international conference on Multimedia}, 2018, pp. 1455--1456.

\bibitem{LeNet5}
Y.~LeCun, L.~Bottou, Y.~Bengio, and P.~Haffner, ``Gradient-based learning
  applied to document recognition,'' \emph{Proceedings of the IEEE}, vol.~86,
  no.~11, pp. 2278--2324, 1998.

\bibitem{ji20123d}
S.~Ji, W.~Xu, M.~Yang, and K.~Yu, ``3d convolutional neural networks for human
  action recognition,'' \emph{IEEE transactions on pattern analysis and machine
  intelligence}, vol.~35, no.~1, pp. 221--231, 2012.

\bibitem{chen20193d}
Y.-C. Chen, D.~S. Tan, W.-H. Cheng, and K.-L. Hua, ``3d object completion via
  class-conditional generative adversarial network,'' in \emph{International
  Conference on Multimedia Modeling}.\hskip 1em plus 0.5em minus 0.4em\relax
  Springer, 2019, pp. 54--66.

\bibitem{hua2015computer_20}
K.-L. Hua, C.-H. Hsu, S.~C. Hidayati, W.-H. Cheng, and Y.-J. Chen,
  ``Computer-aided classification of lung nodules on computed tomography images
  via deep learning technique,'' \emph{OncoTargets and therapy}, vol.~8, 2015.

\bibitem{peng2017dual}
M.~Peng, C.~Wang, T.~Chen, G.~Liu, and X.~Fu, ``Dual temporal scale
  convolutional neural network for micro-expression recognition,''
  \emph{Frontiers in Psychology}, vol.~8, p. 1745, 2017.

\bibitem{takalkar2017image}
M.~A. Takalkar and M.~Xu, ``Image based facial micro-expression recognition
  using deep learning on small datasets,'' in \emph{IEEE International
  Conference on Digital Image Computing: Techniques and Applications}.\hskip
  1em plus 0.5em minus 0.4em\relax IEEE, 2017, pp. 1--7.

\bibitem{wu2020tsnn}
C.~Wu and F.~Guo, ``Tsnn: Three-stream combining 2d and 3d convolutional neural
  network for micro-expression recognition,'' \emph{IEEJ Transactions on
  Electrical and Electronic Engineering}, 2020.

\bibitem{khor2018enriched}
H.-Q. Khor, J.~See, R.~C.~W. Phan, and W.~Lin, ``Enriched long-term recurrent
  convolutional network for facial micro-expression recognition,'' in
  \emph{IEEE International Conference on Automatic Face and Gesture
  Recognition}.\hskip 1em plus 0.5em minus 0.4em\relax IEEE, 2018, pp.
  667--674.

\bibitem{li2019micro}
J.~Li, Y.~Wang, J.~See, and W.~Liu, ``Micro-expression recognition based on 3d
  flow convolutional neural network,'' \emph{Pattern Analysis and
  Applications}, vol.~22, no.~4, pp. 1331--1339, 2019.

\bibitem{reddy2019spontaneous}
S.~P.~T. Reddy, S.~T. Karri, S.~R. Dubey, and S.~Mukherjee, ``Spontaneous
  facial micro-expression recognition using 3d spatiotemporal convolutional
  neural networks,'' in \emph{IEEE International Joint Conference on Neural
  Networks}.\hskip 1em plus 0.5em minus 0.4em\relax IEEE, 2019, pp. 1--8.

\bibitem{gupta2018exploring}
P.~Gupta, B.~Bhowmick, and A.~Pal, ``Exploring the feasibility of face video
  based instantaneous heart-rate for micro-expression spotting,'' in \emph{IEEE
  International Conference on Computer Vision and Pattern Recognition
  workshops}.\hskip 1em plus 0.5em minus 0.4em\relax IEEE, 2018, pp.
  1316--1323.

\bibitem{tran2020micro}
T.-K. Tran, Q.-N. Vo, X.~Hong, X.~Li, and G.~Zhao, ``Micro-expression spotting:
  A new benchmark,'' \emph{arXiv preprint arXiv:2007.12421}, 2020.

\bibitem{shreve2009towards}
M.~Shreve, S.~Godavarthy, V.~Manohar, D.~Goldgof, and S.~Sarkar, ``Towards
  macro-and micro-expression spotting in video using strain patterns,'' in
  \emph{IEEE Workshop on Applications of Computer Vision}.\hskip 1em plus 0.5em
  minus 0.4em\relax IEEE, 2009, pp. 1--6.

\bibitem{shreve2011macro}
M.~Shreve, S.~Godavarthy, D.~Goldgof, and S.~Sarkar, ``Macro-and
  micro-expression spotting in long videos using spatio-temporal strain,'' in
  \emph{IEEE International Conference on Automatic Face and Gesture
  Recognition}.\hskip 1em plus 0.5em minus 0.4em\relax IEEE, 2011, pp. 51--56.

\bibitem{zhang2018smeconvnet}
Z.~Zhang, T.~Chen, H.~Meng, G.~Liu, and X.~Fu, ``Smeconvnet: A convolutional
  neural network for spotting spontaneous facial micro-expression from long
  videos,'' \emph{IEEE Access}, vol.~6, pp. 71\,143--71\,151, 2018.

\bibitem{liong2015automaticapex}
S.-T. Liong, J.~See, K.~Wong, A.~C. Le~Ngo, Y.-H. Oh, and R.~Phan, ``Automatic
  apex frame spotting in micro-expression database,'' in \emph{IEEE Asian
  conference on pattern recognition}.\hskip 1em plus 0.5em minus 0.4em\relax
  IEEE, 2015, pp. 665--669.

\bibitem{FERSurvey2019}
S.~Li and W.~Deng, ``Deep facial expression recognition: A survey,'' \emph{IEEE
  Transactions on Affective Computing}, 2020.

\bibitem{weber2018survey}
R.~Weber, J.~Li, C.~Soladie, and R.~Seguier, ``A survey on databases of facial
  macro-expression and micro-expression,'' in \emph{International Joint
  Conference on Computer Vision, Imaging and Computer Graphics}.\hskip 1em plus
  0.5em minus 0.4em\relax Springer, 2018, pp. 298--325.

\bibitem{tan2018survey}
C.~Tan, F.~Sun, T.~Kong, W.~Zhang, C.~Yang, and C.~Liu, ``A survey on deep
  transfer learning,'' in \emph{International conference on artificial neural
  networks}.\hskip 1em plus 0.5em minus 0.4em\relax Springer, 2018, pp.
  270--279.

\bibitem{zhao2011facial}
G.~Zhao, X.~Huang, M.~Taini, S.~Z. Li, and M.~Pietik{\"a}Inen, ``Facial
  expression recognition from near-infrared videos,'' \emph{Image and Vision
  Computing}, vol.~29, no.~9, pp. 607--619, 2011.

\bibitem{lucey2010extended}
P.~Lucey, J.~F. Cohn, T.~Kanade, J.~Saragih, Z.~Ambadar, and I.~Matthews, ``The
  extended cohn-kanade dataset (ck+): A complete dataset for action unit and
  emotion-specified expression,'' in \emph{IEEE Computer Society Conference on
  Computer Vision and Pattern Recognition-Workshops}.\hskip 1em plus 0.5em
  minus 0.4em\relax IEEE, 2010, pp. 94--101.

\bibitem{lyons1998coding}
M.~Lyons, S.~Akamatsu, M.~Kamachi, and J.~Gyoba, ``Coding facial expressions
  with gabor wavelets,'' in \emph{IEEE international conference on automatic
  face and gesture recognition}.\hskip 1em plus 0.5em minus 0.4em\relax IEEE,
  1998, pp. 200--205.

\bibitem{aifanti2010mug}
N.~Aifanti, C.~Papachristou, and A.~Delopoulos, ``The mug facial expression
  database,'' in \emph{11th International Workshop on Image Analysis for
  Multimedia Interactive Services WIAMIS 10}.\hskip 1em plus 0.5em minus
  0.4em\relax IEEE, 2010, pp. 1--4.

\bibitem{valstar2017fera}
M.~F. Valstar, E.~S{\'a}nchez-Lozano, J.~F. Cohn, L.~A. Jeni, J.~M. Girard,
  Z.~Zhang, L.~Yin, and M.~Pantic, ``Fera 2017-addressing head pose in the
  third facial expression recognition and analysis challenge,'' in \emph{IEEE
  international conference on automatic face and gesture recognition}.\hskip
  1em plus 0.5em minus 0.4em\relax IEEE, 2017, pp. 839--847.

\bibitem{zhu2017unpaired}
J.-Y. Zhu, T.~Park, P.~Isola, and A.~A. Efros, ``Unpaired image-to-image
  translation using cycle-consistent adversarial networks,'' in
  \emph{Proceedings of the IEEE international conference on computer vision},
  2017, pp. 2223--2232.

\bibitem{Allaert2019lmp}
B.~Allaert, I.~M. Bilasco, and C.~Djeraba, ``Micro and macro facial expression
  recognition using advanced local motion patterns,'' \emph{IEEE Transactions
  on Affective Computing}, 2019.

\bibitem{asmara2019drmf}
R.~Asmara, P.~Choirina, C.~Rahmad, A.~Setiawan, F.~Rahutomo, R.~Yusron, and
  U.~Rosiani, ``Study of drmf and asm facial landmark point for micro
  expression recognition using klt tracking point feature,'' in \emph{Journal
  of Physics: Conference Series}, vol. 1402, no.~7.\hskip 1em plus 0.5em minus
  0.4em\relax IOP Publishing, 2019, p. 077039.

\bibitem{lu2018motion}
H.~Lu, K.~Kpalma, and J.~Ronsin, ``Motion descriptors for micro-expression
  recognition,'' \emph{Signal Processing: Image Communication}, vol.~67, pp.
  108--117, 2018.

\bibitem{Zhou2019dual}
L.~Zhou, Q.~Mao, and L.~Xue, ``Dual-inception network for cross-database
  micro-expression recognition,'' in \emph{IEEE International Conference on
  Automatic Face and Gesture Recognition}.\hskip 1em plus 0.5em minus
  0.4em\relax IEEE, 2019, pp. 1--5.

\bibitem{verma2019learnet}
M.~Verma, S.~K. Vipparthi, G.~Singh, and S.~Murala, ``Learnet: Dynamic imaging
  network for micro expression recognition,'' \emph{IEEE Transactions on Image
  Processing}, vol.~29, pp. 1618--1627, 2019.

\bibitem{xia2019spatiotemporal}
Z.~Xia, X.~Hong, X.~Gao, X.~Feng, and G.~Zhao, ``Spatiotemporal recurrent
  convolutional networks for recognizing spontaneous micro-expressions,''
  \emph{IEEE Transactions on Multimedia}, vol.~22, no.~3, pp. 626--640, 2019.

\bibitem{xia2019cross}
Z.~Xia, H.~Liang, X.~Hong, and X.~Feng, ``Cross-database micro-expression
  recognition with deep convolutional networks,'' in \emph{ACM International
  Conference on Biometric Engineering and Applications}.\hskip 1em plus 0.5em
  minus 0.4em\relax ACM, 2019, pp. 56--60.

\bibitem{luself}
L.~Lu, L.~Tavabi, and M.~Soleymani, ``Self-supervised learning for facial
  action unit recognition through temporal consistency,'' in \emph{British
  Machine Vision Conference}.\hskip 1em plus 0.5em minus 0.4em\relax BMVA,
  2020.

\bibitem{wang2019multi}
C.~Wang, J.~Zeng, S.~Shan, and X.~Chen, ``Multi-task learning of emotion
  recognition and facial action unit detection with adaptively weights sharing
  network,'' in \emph{IEEE International Conference on Image Processing}.\hskip
  1em plus 0.5em minus 0.4em\relax IEEE, 2019, pp. 56--60.

\bibitem{aurcnn}
C.~Ma, L.~Chen, and J.~Yong, ``Au r-cnn: Encoding expert prior knowledge into
  r-cnn for action unit detection,'' \emph{Neurocomputing}, vol. 355, pp.
  35--47, 2019.

\bibitem{aucrossdomain}
I.~O. Ertugrul, J.~F. Cohn, L.~A. Jeni, Z.~Zhang, L.~Yin, and Q.~Ji,
  ``Cross-domain au detection: Domains, learning approaches, and measures,'' in
  \emph{IEEE International Conference on Automatic Face and Gesture
  Recognition}, 2019, pp. 1--8.

\bibitem{aumultilabel}
W.~Li, F.~Abtahi, and Z.~Zhu, ``Action unit detection with region adaptation,
  multi-labeling learning and optimal temporal fusing,'' in \emph{IEEE
  International Conference on Computer Vision and Pattern Recognition}.\hskip
  1em plus 0.5em minus 0.4em\relax IEEE, 2017, pp. 1841--1850.

\bibitem{eacnet}
W.~Li, F.~Abtahi, Z.~Zhu, and L.~Yin, ``Eac-net: A region-based deep enhancing
  and cropping approach for facial action unit detection,'' in \emph{IEEE
  International Conference on Automatic Face and Gesture Recognition}.\hskip
  1em plus 0.5em minus 0.4em\relax IEEE, 2017, pp. 103--110.

\bibitem{fan2020AUintensity}
Y.~Fan, J.~C. Lam, and V.~O.~K. Li, ``Facial action unit intensity estimation
  via semantic correspondence learning with dynamic graph convolution.'' in
  \emph{AAAI Conference on Artificial Intelligence}.\hskip 1em plus 0.5em minus
  0.4em\relax AAAI, 2020, pp. 12\,701--12\,708.

\bibitem{li2020learning}
Y.~Li, J.~Zeng, and S.~Shan, ``Learning representations for facial actions from
  unlabeled videos,'' \emph{IEEE Transactions on Pattern Analysis and Machine
  Intelligence}, 2020.

\bibitem{zhi2020_AUSurvey}
R.~Zhi, M.~Liu, and D.~Zhang, ``A comprehensive survey on automatic facial
  action unit analysis,'' \emph{The Visual Computer}, vol.~36, no.~5, pp.
  1067--1093, 2020.

\bibitem{martinez2017_AUSurvey}
B.~Martinez, M.~F. Valstar, B.~Jiang, and M.~Pantic, ``Automatic analysis of
  facial actions: A survey,'' \emph{IEEE Transactions on Affective Computing},
  vol.~10, no.~3, pp. 325--347, 2019.

\bibitem{liu2020relation}
Z.~Liu, J.~Dong, C.~Zhang, L.~Wang, and J.~Dang, ``Relation modeling with graph
  convolutional networks for facial action unit detection,'' in
  \emph{International Conference on Multimedia Modeling}.\hskip 1em plus 0.5em
  minus 0.4em\relax Springer, 2020, pp. 489--501.

\bibitem{li2019semantic}
G.~Li, X.~Zhu, Y.~Zeng, Q.~Wang, and L.~Lin, ``Semantic relationships guided
  representation learning for facial action unit recognition,'' in \emph{AAAI
  Conference on Artificial Intelligence}, vol.~33.\hskip 1em plus 0.5em minus
  0.4em\relax AAAI, 2019, pp. 8594--8601.

\bibitem{chen2019multi}
Z.-M. Chen, X.-S. Wei, P.~Wang, and Y.~Guo, ``Multi-label image recognition
  with graph convolutional networks,'' in \emph{IEEE International Conference
  on Computer Vision and Pattern Recognition}.\hskip 1em plus 0.5em minus
  0.4em\relax IEEE, 2019, pp. 5177--5186.

\bibitem{autraditional}
W.-S. Chu, F.~De~la Torre, and J.~F. Cohn, ``Learning spatial and temporal cues
  for multi-label facial action unit detection,'' in \emph{IEEE International
  Conference on Automatic Face and Gesture Recognition}.\hskip 1em plus 0.5em
  minus 0.4em\relax IEEE, 2017, pp. 25--32.

\bibitem{autraditional2}
K.~Zhao, W.-S. Chu, F.~De~la Torre, J.~F. Cohn, and H.~Zhang, ``Joint patch and
  multi-label learning for facial action unit and holistic expression
  recognition,'' \emph{IEEE Transactions on Image Processing}, vol.~25, no.~8,
  pp. 3931--3946, 2016.

\bibitem{liu2015main}
Y.-J. Liu, J.-K. Zhang, W.-J. Yan, S.-J. Wang, G.~Zhao, and X.~Fu, ``A main
  directional mean optical flow feature for spontaneous micro-expression
  recognition,'' \emph{IEEE Transactions on Affective Computing}, vol.~7,
  no.~4, pp. 299--310, 2015.

\bibitem{lo2020mer-gcn}
L.~Lo, H.-X. Xie, H.-H. Shuai, and W.-H. Cheng, ``Mer-gcn: Micro-expression
  recognition based on relation modeling with graph convolutional networks,''
  in \emph{IEEE Conference on Multimedia Information Processing and
  Retrieval}.\hskip 1em plus 0.5em minus 0.4em\relax IEEE, 2020, pp. 79--84.

\bibitem{li2020deep}
Q.~Li, J.~Yu, T.~Kurihara, H.~Zhang, and S.~Zhan, ``Deep convolutional neural
  network with optical flow for facial micro-expression recognition,''
  \emph{Journal of Circuits, Systems and Computers}, vol.~29, no.~01, p.
  2050006, 2020.

\bibitem{merghani2020adaptive}
W.~Merghani and M.~H. Yap, ``Adaptive mask for region-based facial
  micro-expression recognition,'' in \emph{IEEE International Conference on
  Automatic Face and Gesture Recognition}.\hskip 1em plus 0.5em minus
  0.4em\relax IEEE, 2020, pp. 428--433.

\bibitem{takalkar2020manifold}
M.~A. Takalkar, M.~Xu, and Z.~Chaczko, ``Manifold feature integration for
  micro-expression recognition,'' \emph{Multimedia Systems}, vol.~26, no.~5,
  pp. 535--551, 2020.

\bibitem{lim2017fuzzy}
C.~H. Lim and K.~M. Goh, ``Fuzzy qualitative approach for micro-expression
  recognition,'' in \emph{Asia-Pacific Signal and Information Processing
  Association Annual Summit and Conference}.\hskip 1em plus 0.5em minus
  0.4em\relax IEEE, 2017, pp. 1669--1674.

\bibitem{pan2019furniture}
T.-Y. Pan, Y.-Z. Dai, M.-C. Hu, and W.-H. Cheng, ``Furniture style
  compatibility recommendation with cross-class triplet loss,''
  \emph{Multimedia Tools and Applications}, vol.~78, no.~3, pp. 2645--2665,
  2019.

\bibitem{smola2004tutorial}
A.~J. Smola and B.~Sch{\"o}lkopf, ``A tutorial on support vector regression,''
  \emph{Statistics and computing}, vol.~14, no.~3, pp. 199--222, 2004.

\bibitem{taigman2014deepface}
Y.~Taigman, M.~Yang, M.~Ranzato, and L.~Wolf, ``Deepface: Closing the gap to
  human-level performance in face verification,'' in \emph{IEEE International
  Conference on Computer Vision and Pattern Recognition}.\hskip 1em plus 0.5em
  minus 0.4em\relax IEEE, 2014, pp. 1701--1708.

\bibitem{lalitha2020micro}
S.~Lalitha and K.~Thyagharajan, ``Micro-facial expression recognition based on
  deep-rooted learning algorithm,'' \emph{International Journal of
  Computational Intelligence Systems}, vol.~12, pp. 903--913, 2019.

\bibitem{lin2017focal}
T.-Y. Lin, P.~Goyal, R.~Girshick, K.~He, and P.~Doll{\'a}r, ``Focal loss for
  dense object detection,'' in \emph{IEEE International Conference on Computer
  Vision}.\hskip 1em plus 0.5em minus 0.4em\relax IEEE, 2017, pp. 2980--2988.

\bibitem{lai2020real}
Z.~Lai, R.~Chen, J.~Jia, and Y.~Qian, ``Real-time micro-expression recognition
  based on resnet and atrous convolutions,'' \emph{Journal of Ambient
  Intelligence and Humanized Computing}, pp. 1--12, 2020.

\bibitem{xie2019adaptive}
W.~Xie, L.~Shen, and J.~Duan, ``Adaptive weighting of handcrafted feature
  losses for facial expression recognition,'' \emph{IEEE Transactions on
  Cybernetics}, 2019.

\bibitem{zhang2020cross}
T.~Zhang, Y.~Zong, W.~Zheng, C.~P. Chen, X.~Hong, C.~Tang, Z.~Cui, and G.~Zhao,
  ``Cross-database micro-expression recognition: A benchmark,'' \emph{IEEE
  Transactions on Knowledge and Data Engineering}, 2020.

\bibitem{zhang2020multiple}
X.~Zhang, T.~Xu, W.~Sun, and A.~Song, ``Multiple source domain adaptation in
  micro-expression recognition,'' \emph{Journal of Ambient Intelligence and
  Humanized Computing}, pp. 1--16, 2020.

\bibitem{li2019talsr}
L.~Li, X.~Zhou, Y.~Zong, W.~Zheng, X.~Chen, J.~Shi, and P.~Song, ``Unsupervised
  cross-database micro-expression recognition using target-adapted
  least-squares regression,'' \emph{IEICE Transactions on Information and
  Systems}, vol. 102, no.~7, pp. 1417--1421, 2019.

\bibitem{wilson2020survey}
G.~Wilson and D.~J. Cook, ``A survey of unsupervised deep domain adaptation,''
  \emph{ACM Transactions on Intelligent Systems and Technology}, vol.~11,
  no.~5, pp. 1--46, 2020.

\bibitem{zhang2019transfer}
L.~Zhang, ``Transfer adaptation learning: A decade survey,'' \emph{arXiv
  preprint arXiv:1903.04687}, 2019.

\bibitem{kumar2019classification}
A.~J.~R. Kumar, R.~Theagarajan, O.~Peraza, and B.~Bhanu, ``Classification of
  facial micro-expressions using motion magnified emotion avatar images.'' in
  \emph{IEEE International Conference on Computer Vision and Pattern
  Recognition Workshops}.\hskip 1em plus 0.5em minus 0.4em\relax IEEE, 2019,
  pp. 12--20.

\bibitem{zadeh1965fuzzy}
L.~A. Zadeh, ``Fuzzy sets,'' \emph{Information and Control}, vol.~8, no.~3, pp.
  338--353, 1965.

\bibitem{chen2016emotion}
M.~Chen, H.~T. Ma, J.~Li, and H.~Wang, ``Emotion recognition using fixed length
  micro-expressions sequence and weighting method,'' in \emph{IEEE
  International Conference on Real-time Computing and Robotics}.\hskip 1em plus
  0.5em minus 0.4em\relax IEEE, 2016, pp. 427--430.

\bibitem{elfenbein2002there}
H.~A. Elfenbein and N.~Ambady, ``Is there an in-group advantage in emotion
  recognition?'' \emph{Psychological Bulletin}, vol. 128, pp. 243--9, 2002.

\bibitem{young2018minimal}
S.~G. Young and J.~P. Wilson, ``A minimal ingroup advantage in emotion
  identification confidence,'' \emph{Cognition and Emotion}, vol.~32, no.~1,
  pp. 192--199, 2018.

\bibitem{tuminello2011face}
E.~R. Tuminello and D.~Davidson, ``What the face and body reveal: In-group
  emotion effects and stereotyping of emotion in african american and european
  american children,'' \emph{Journal of Experimental Child Psychology}, vol.
  110, no.~2, pp. 258--274, 2011.

\bibitem{thibault2006effect}
P.~Thibault, P.~Bourgeois, and U.~Hess, ``The effect of group-identification on
  emotion recognition: The case of cats and basketball players,'' \emph{Journal
  of Experimental Social Psychology}, vol.~42, no.~5, pp. 676--683, 2006.

\bibitem{young2010mere}
S.~G. Young and K.~Hugenberg, ``Mere social categorization modulates
  identification of facial expressions of emotion.'' \emph{Journal of
  Personality and Social Psychology}, vol.~99, no.~6, p. 964, 2010.

\bibitem{huang2014shared}
S.~Huang and S.~Han, ``Shared beliefs enhance shared feelings:
  religious/irreligious identifications modulate empathic neural responses,''
  \emph{Social Neuroscience}, vol.~9, no.~6, pp. 639--649, 2014.

\bibitem{prado2014facial}
C.~Prado, D.~Mellor, L.~K. Byrne, C.~Wilson, X.~Xu, and H.~Liu, ``Facial
  emotion recognition: a cross-cultural comparison of chinese, chinese living
  in australia, and anglo-australians,'' \emph{Motivation and Emotion},
  vol.~38, no.~3, pp. 420--428, 2014.

\bibitem{xie2019ingroup}
Y.~Xie, C.~Zhong, F.~Zhang, and Q.~Wu, ``The ingroup disadvantage in the
  recognition of micro-expressions,'' in \emph{IEEE International Conference on
  Automatic Face and Gesture Recognition}.\hskip 1em plus 0.5em minus
  0.4em\relax IEEE, 2019, pp. 1--5.

\bibitem{folster2014facial}
M.~F{\"o}lster, U.~Hess, and K.~Werheid, ``Facial age affects emotional
  expression decoding,'' \emph{Frontiers in psychology}, vol.~5, p.~30, 2014.

\bibitem{chen2020label}
S.~Chen, J.~Wang, Y.~Chen, Z.~Shi, X.~Geng, and Y.~Rui, ``Label distribution
  learning on auxiliary label space graphs for facial expression recognition,''
  in \emph{Proceedings of the IEEE/CVF Conference on Computer Vision and
  Pattern Recognition}, 2020, pp. 13\,984--13\,993.

\bibitem{wang2020suppressing}
K.~Wang, X.~Peng, J.~Yang, S.~Lu, and Y.~Qiao, ``Suppressing uncertainties for
  large-scale facial expression recognition,'' in \emph{Proceedings of the
  IEEE/CVF Conference on Computer Vision and Pattern Recognition}, 2020, pp.
  6897--6906.

\bibitem{mandal1995asymmetry}
M.~K. Mandal, H.~S. Asthana, and R.~Pandey, ``Asymmetry in emotional face: Its
  role in intensity of expression,'' \emph{The Journal of Psychology}, vol.
  129, no.~2, pp. 235--241, 1995.

\bibitem{dopson1984asymmetry}
W.~G. Dopson, B.~E. Beckwith, D.~M. Tucker, and P.~C. Bullard-Bates,
  ``Asymmetry of facial expression in spontaneous emotion,'' \emph{Cortex},
  vol.~20, no.~2, pp. 243--251, 1984.

\bibitem{chen2019analyze}
H.~Chen, X.~Liu, X.~Li, H.~Shi, and G.~Zhao, ``Analyze spontaneous gestures for
  emotional stress state recognition: A micro-gesture dataset and analysis with
  deep learning,'' in \emph{IEEE International Conference on Automatic Face and
  Gesture Recognition}.\hskip 1em plus 0.5em minus 0.4em\relax IEEE, 2019, pp.
  1--8.

\end{thebibliography}
%



%

\begin{IEEEbiography}[{\includegraphics[width=1in,height=1.25in,clip,keepaspectratio]{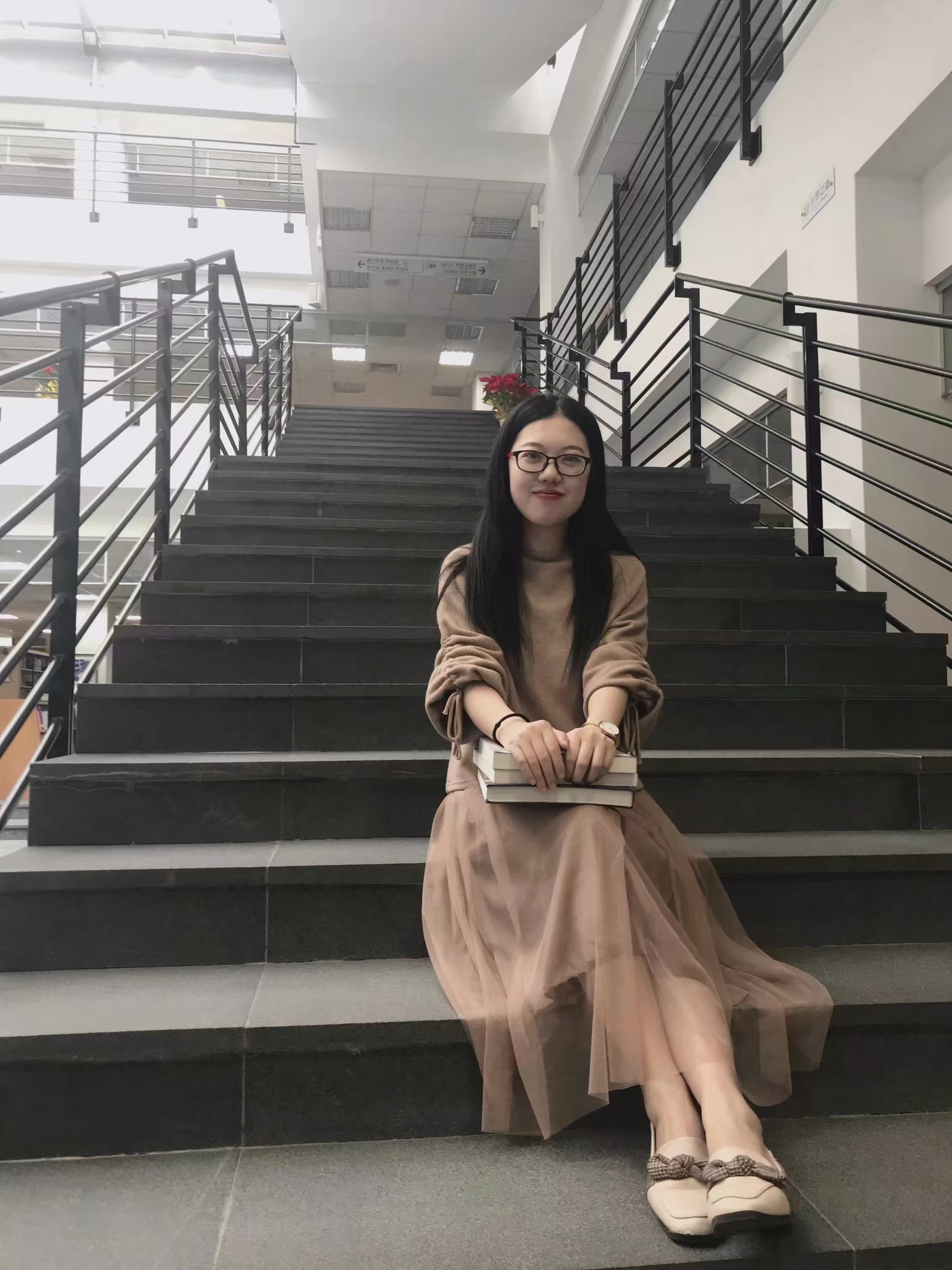}}]{Hong-Xia Xie}
received the B.S. degree in Internet of Things from the Zhengzhou University of Aeronautics in 2016 and received the M.S. degree in communication and information systems, Fujian Normal
University, China, in 2019. She is now pursuing a Ph.D. degree in Institute of of Electronics, National Chiao Tung Univresity, Taiwan. Her research interests include micro-expression recognition, emotion recognition and deep learning.
\end{IEEEbiography}

\begin{IEEEbiography}[{\includegraphics[width=1in,height=1.25in,clip,keepaspectratio]{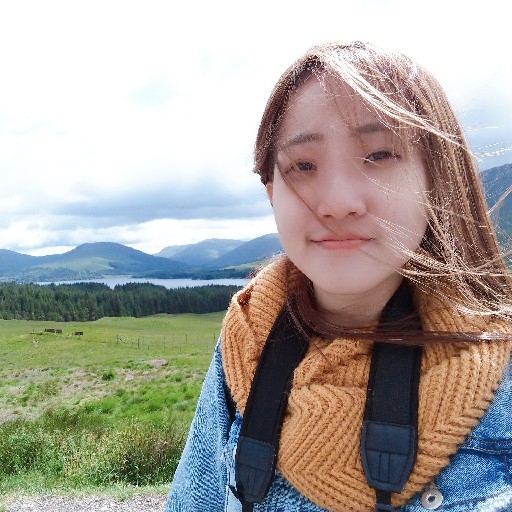}}]{Ling Lo}
received the B.S. degree from Department of Electronics Engineering, National Chiao Tung University (NCTU), Hsinchu, Taiwan, R.O.C., in 2019, and now she is pursuing a Ph.D degree in Institute of Electronics, NCTU. Her current research interests include deep learning and computer vision. Recently her work focuses specifically on facial and micro-expression recognition.
\end{IEEEbiography}

\begin{IEEEbiography}[{\includegraphics[width=1in,height=1.25in,clip,keepaspectratio]{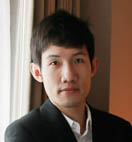}}]{Hong-Han Shuai} received the B.S. degree from the Department of Electrical Engineering, National Taiwan University (NTU), Taipei, Taiwan, R.O.C., in 2007, the M.S. degree in computer science from NTU in 2009, and the Ph.D. degree from Graduate Institute of Communication Engineering, NTU, in 2015. He is now an assistant professor in NCTU. His research interests are in the area of multimedia processing, machine learning, social network analysis, and data mining. His works have appeared in top-tier conferences such as MM, CVPR, AAAI, KDD, WWW, ICDM, CIKM and VLDB, and top-tier journals such as TKDE, TMM and JIOT. Moreover, he has served as the PC member for international conferences including MM, AAAI, IJCAI, WWW, and the invited reviewer for journals including TKDE, TMM, JVCI and JIOT.
\end{IEEEbiography}

\begin{IEEEbiography}[{\includegraphics[width=1in,clip,keepaspectratio]{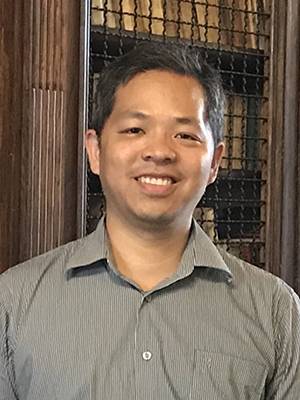}}]{Wen-Huang Cheng} is Professor with the Institute of Electronics, National Chiao Tung University (NCTU), Hsinchu, Taiwan. He is also Jointly Appointed Professor with the Artificial  Intelligence and Data Science Program, National Chung Hsing University (NCHU), Taichung, Taiwan. Before joining NCTU, he led the Multimedia Computing Research Group at the Research Center for Information Technology Innovation (CITI), Academia Sinica, Taipei, Taiwan, from 2010 to 2018. His current research interests include multimedia, artificial intelligence, computer vision, and machine learning. He has actively participated in international events and played important leading roles in prestigious journals and conferences and professional organizations, like Associate Editor for IEEE Transactions on Multimedia, General co-chair for IEEE ICME (2022) and ACM ICMR (2021), Chair-Elect for IEEE MSA technical committee, governing board member for IAPR. He has received numerous research and service awards, including the 2018 MSRA Collaborative Research Award, the 2017 Ta-Yu Wu Memorial Award from Taiwan’s Ministry of Science and Technology (the highest national research honor for young Taiwanese researchers under age 42), the 2017 Significant Research Achievements of Academia Sinica, the Top 10\% Paper Award from the 2015 IEEE MMSP, and the K. T. Li Young Researcher Award from the ACM Taipei/Taiwan Chapter in 2014. He is IET Fellow and ACM Distinguished Member.
\end{IEEEbiography}




\end{document}